\documentclass[a4paper,fleqn]{cas-dc}
\usepackage[numbers,sort&compress]{natbib}
\usepackage{amsmath}
\usepackage[ruled,vlined,linesnumbered]{algorithm2e}
\DontPrintSemicolon
\usepackage{graphicx}
\usepackage{xcolor}
\usepackage{colortbl}
\usepackage{float}
\usepackage{array}
\usepackage{seqsplit}
\usepackage{ragged2e}
\usepackage{placeins}
\hypersetup{hypertexnames=false}

\makeatletter
\renewcommand{\NAT@def@citea}{%
  \def\@citea{\NAT@@close\NAT@separator\NAT@space\NAT@@open}%
}
\patchcmd{\NAT@citexnum}
  {\def@NAT@last@yr{--\NAT@penalty}}
  {\def@NAT@last@yr{\NAT@@close-\NAT@penalty\NAT@@open}}
  {}{\PackageError{cuedrnet}{Unable to configure citation ranges}{}}
\makeatother

\makeatletter
\def\@dbflt#1{%
  \@ifnextchar[{\@xdblfloat{#1}}{%
    \edef\reserved@b{\csname fps@#1\endcsname}%
    \ifx\reserved@b\@empty\def\reserved@b{tp}\fi
    \edef\reserved@a{\noexpand\@xdblfloat{#1}[\reserved@b]}%
    \reserved@a
  }%
}
\makeatother

\newcommand{\emailaddr}[1]{%
  {\ttfamily\seqsplit{#1}}%
}

\renewcommand{\printemails}{%
  \begingroup
  \let\thefootnote\relax
  \footnotetext{\justifying\scriptsize\sloppy
    \includegraphics[height=8pt]{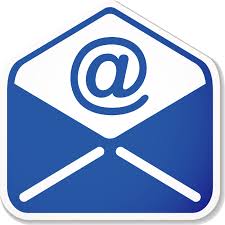}\space
    \emailaddr{yicuishi@cqu.edu.cn} (Y.\ Shi);
    \emailaddr{20240701028@stu.cqu.edu.cn} (Y.\ Chen);
    \emailaddr{hxf0213@stu.cqu.edu.cn} (X.\ Huang);
    \emailaddr{wangzhenguo\_ssz@zpmc.com} (Z.\ Wang);
    \emailaddr{wenxuanyu@cqu.edu.cn} (W.\ Yu);
    \emailaddr{yingfang@stu.cqu.edu.cn} (Y.\ Fang);
    \emailaddr{jieli@cqu.edu.cn} (J.\ Li);
    \emailaddr{liguofa@cqu.edu.cn} (G.\ Li)%
  }%
  \endgroup
}

\definecolor{bestcol}{HTML}{F16913}
\definecolor{secondcol}{HTML}{FDD0A2}
\newcommand{\best}[1]{\cellcolor{bestcol}#1}
\newcommand{\seco}[1]{\cellcolor{secondcol}#1}

\def\tsc#1{\csdef{#1}{\textsc{\lowercase{#1}}\xspace}}
\tsc{WGM}
\tsc{QE}

\begin{document}
\let\WriteBookmarks\relax
\def\floatpagepagefraction{1}
\def\textpagefraction{.001}

\shorttitle{CueDR-Net: Cue-Guided Dual-Path Reconstruction}

\shortauthors{Y. Shi et al.}

\title [mode = title]{Cue-guided dual-path reconstruction with adaptive multi-scale response refinement for low-light image enhancement}

\author[1]{Yicui Shi}
\ead{yicuishi@cqu.edu.cn}
\author[1]{Yuhan Chen}
\ead{20240701028@stu.cqu.edu.cn}
\author[1]{Xiangfei Huang}
\ead{hxf0213@stu.cqu.edu.cn}
\author[2]{Zhenguo Wang}
\ead{wangzhenguo_ssz@zpmc.com}
\author[1]{Wenxuan Yu}
\ead{wenxuanyu@cqu.edu.cn}
\author[1]{Ying Fang}
\ead{yingfang@stu.cqu.edu.cn}
\author[1]{Jie Li}\cormark[1]
\ead{jieli@cqu.edu.cn}
\author[1]{Guofa Li}[orcid=0000-0002-7889-4695]
\ead{liguofa@cqu.edu.cn} 
\affiliation[1]{organization={College of Mechanical and Vehicle Engineering, Chongqing University},
            city={Chongqing},
            postcode={400044},
            country={China}}
\affiliation[2]{organization={Shanghai Zhenhua Heavy Industries Co.,Ltd.},
            city={Shanghai},
            postcode={200125},
            country={China}}
\cortext[1]{Corresponding author}

\begin{abstract}
Low-light image enhancement (LLIE) supports optical sensing and scene observation when limited photon capture and sensor noise degrade image quality. However, existing methods can introduce colour deviations or blur local details in the enhanced images. To address this challenge, we propose \textbf{CueDR-Net}, a dual-path network that combines brightness adjustment with local content reconstruction through bidirectional cue interaction and adaptive multi-scale refinement. Specifically, a fixed extractor derives luminance variation, colour, and texture cue maps from the input image. These maps are encoded in a cue stream that exchanges features with the image stream through bidirectional attention. The exchanged features support two paths for brightness adjustment and local content reconstruction. The brightness adjustment path predicts pixel wise adjustment maps that control iterative intensity changes in the input image. In parallel, the residual path uses the exchanged features to predict an additive correction for local content reconstruction. To capture weak structures across scales, Adaptive Multi-Scale Response Refinement Convolution (AMRRConv) refines the features used for both predictions. It fuses learnable local transition responses with multi-scale features from factorized three-plane filtering using pixel wise weights. The adjusted image is added to the correction before being mapped to the enhanced image. Evaluations on LOLv1, LSRW-HUAWEI, and LSRW-NIKON demonstrate competitive enhancement performance, supported by ablation studies. AMRRConv also achieves competitive low-light detection performance with a small neck parameter count when integrated into YOLOv11.
\end{abstract}

\begin{highlights}
\item A cue-guided framework integrates luminance, chroma, and texture cues.
\item A bidirectional dual-stream design guides complementary reconstruction paths.
\item A multi-scale response refinement block adaptively fuses candidate features.
\end{highlights}


\maketitle

\section{Introduction}\label{sec:intro}

Colour imaging captures scene intensity, colour, and spatial detail for environmental perception and object recognition. Under limited illumination, optical sensors face a trade-off between the low signal-to-noise ratio (SNR) of short exposures and the risk of motion blur during longer exposures \cite{chen2018learning}. Photon shot noise and sensor read noise degrade weak measurements, while demosaicing and colour processing of noisy Bayer measurements can further compromise colour fidelity and fine spatial detail \cite{brooks2019unprocessing}. LLIE uses computational processing to address these degradations, but simple intensity amplification can magnify noise and colour bias. Effective enhancement therefore requires coordinated photometric recovery, colour consistency, and structural preservation \cite{zhao2025deep,li2025comprehensive,sun2025low,ding2025low}.

\begin{figure*}[pos=t]
\centering
\includegraphics[width=\textwidth]{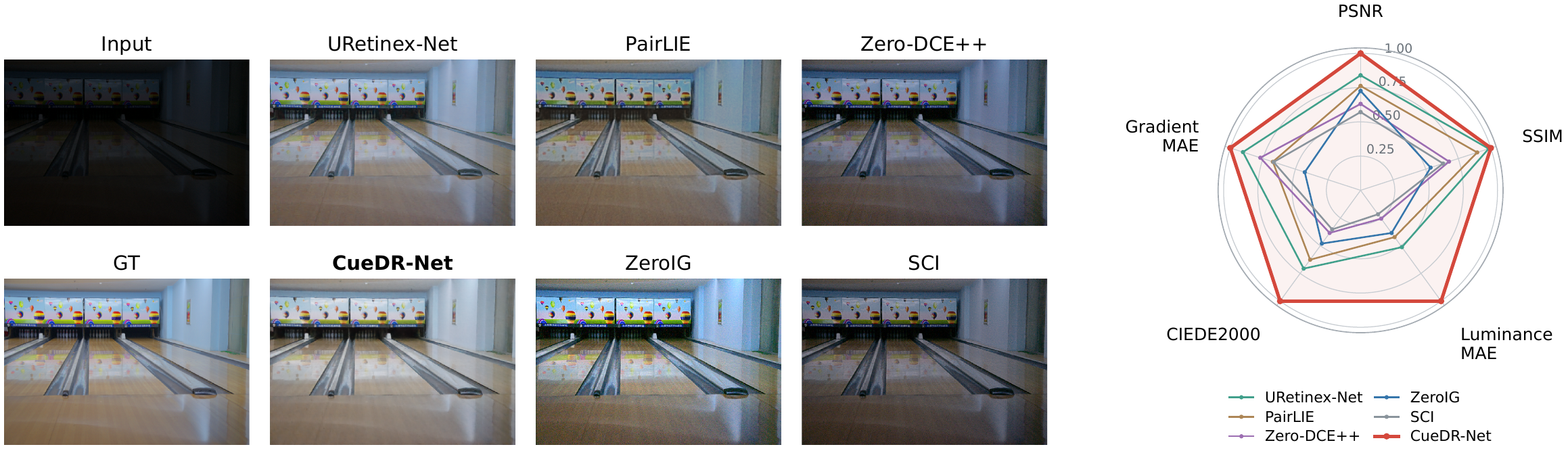}
\caption{Enhancement results on a LOLv1 example (left) and normalized full-test scores (right). The radar summarizes PSNR, SSIM, luminance MAE, CIEDE2000, and gradient MAE. Each metric is displayed on a normalized $[0,1]$ scale, with higher PSNR and SSIM and lower errors corresponding to larger radii.}\label{fig:ole_problem}
\end{figure*}

Existing methods improve low-light images through explicit photometric modelling and learned reconstruction. Zero-DCE estimates pixel wise curves for spatially varying intensity adjustment \cite{guo2020zero}, while IAGC combines explicit gamma correction with learned image modelling \cite{wang2023iagc}. Retinex-based networks separate illumination adjustment from reflectance restoration \cite{zhang2019kind,wu2022uretinex}. LSLE jointly enhances brightness and saturation, highlighting the role of colour in low-light image quality \cite{wang2024lsle}. These approaches offer different ways to coordinate photometric adjustment and content recovery. Although pixel wise adjustment maps can be predicted from spatial features, the brightness adjustment path forms its output through a prescribed transformation of the observed pixel intensities. The residual path instead predicts an additive correction from neighbouring features, so local content changes are not confined to this transformation. Combining the two paths retains explicit photometric adjustment while allowing the output to include this additional correction.

Alongside output design, existing methods also explore structural and chromatic guidance for enhancement. SMG guides appearance synthesis with predicted structure \cite{xu2023smg}, DCC-Net combines grayscale reconstruction with colour histograms \cite{zhang2022dcc}, and HVI-CIDNet exchanges intensity and chromatic features within a dedicated colour representation \cite{yan2025hvi}. Edge-guided methods also exploit local transitions to support enhancement \cite{zhu2020eemefn,lu2022gpanet,gong2023melenet}. These studies show that reconstruction depends on how image information is represented and passed through the network. Their designs emphasize structure prediction, colour distributions, or interactions between intensity and chromatic features. We investigate how structural and chromatic guidance can be organized alongside the original image content and carried through feature extraction to both the brightness adjustment path and the residual path.

To connect this guidance with both output operations, we propose CueDR-Net. The network maintains a dedicated cue stream that interacts with image features throughout reconstruction, allowing auxiliary information to support two complementary operations. The brightness adjustment path applies a prescribed intensity transformation, while the residual path provides additional corrections beyond this transformation. To account for the varying spatial extents of local structures, we further introduce Adaptive Multi-Scale Response Refinement Convolution to refine the features used by both paths. Fig.~\ref{fig:ole_problem} presents representative enhancement results and a five-metric evaluation on the LOLv1 test set. Our main contributions are as follows.
\begin{itemize}
\item We propose CueDR-Net to coordinate photometric adjustment with colour preservation and local detail recovery in low-light images. Bidirectional attention between the image and cue streams uses queries from each stream to attend to keys and values from the other, allowing the brightness adjustment path and the residual path to use image content and auxiliary guidance for iterative intensity adjustment and local content correction respectively.
\item We introduce Adaptive Multi-Scale Response Refinement Convolution to refine weak boundaries and fine textures in low-light images. The module captures local transition responses using learnable Sobel and Laplacian initialized filters, while factorized three-plane filtering incorporates complementary multi-scale context. Pixel-wise adaptive fusion then integrates these features according to local image content, supporting both brightness adjustment map prediction and residual~reconstruction.
\item We evaluate CueDR-Net on LOLv1 and LSRW and validate the proposed designs through ablation studies. Integrating AMRRConv into a downstream low-light detection network further demonstrates its ability to deliver competitive performance with a small parameter count.
\end{itemize}

\section{Related Work}\label{sec:related}

\subsection{Learning-Based Low-Light Image Enhancement}

Low-light image enhancement improves image quality under limited illumination by adjusting intensity distributions or reconstructing image content. Traditional methods use histogram adjustment, gamma correction, and illumination estimation to improve visibility \cite{pizer1987adaptive,huang2012efficient,land1977retinex,guo2016lime}. Learning-based methods further estimate enhancement operations and their parameters from data. Zero-DCE and Zero-DCE++ formulate enhancement as pixel wise curve estimation \cite{guo2020zero,li2021learning}, SCI learns illumination adjustment through self-calibration \cite{ma2022toward}, and IAGC combines explicit gamma correction with local-to-global image modelling \cite{wang2023iagc}. These methods give photometric transformations an explicit form, with their capabilities shaped by both parameter estimation and the chosen transformation. Although curve parameters can be predicted from neighbouring features, the curves themselves adjust observed intensities at individual image pixels.

Other methods recover image content through decomposition or direct mapping. RetinexNet learns illumination--reflectance decomposition \cite{wei2018deep}, KinD and KinD++ separate illumination adjustment from reflectance restoration \cite{zhang2019kind,zhang2021beyond}, and URetinex-Net embeds learned restoration in iterative optimization \cite{wu2022uretinex}. Direct mapping methods learn enhanced images from low-light inputs \cite{fu2023learning}. FMR-Net and FRR-Net use residual and multi-scale processing \cite{chen2024fmr,chen2024frr}, while EnlightenGAN learns an enhancement mapping through unpaired adversarial training \cite{jiang2021enlightengan}. Unlike explicit pixel wise transformations, feature-based restoration can use spatial context to recover local content. These studies organize photometric adjustment and content recovery through decomposition, cascaded processing, or joint mappings, determining how the two operations contribute to the final output.

Structure-guided methods use edges and local transitions to support reconstruction. EEMEFN refines multi-exposure fusion with an estimated edge map \cite{zhu2020eemefn}, GPANet incorporates Sobel and Laplacian priors into multi-view analysis \cite{lu2022gpanet}, SMG guides appearance synthesis with predicted structure \cite{xu2023smg}, and MELENet integrates edge information across resolutions \cite{gong2023melenet}. These methods demonstrate the value of explicit structural representations for recovering contours and details in low-light images. They differ in how these representations are extracted and where they enter the network, from input guidance and feature interaction to later refinement. Edge and gradient responses emphasize local transitions, while colour appearance and texture patterns call for complementary representations.

Colour and texture studies provide complementary perspectives. LSLE jointly enhances brightness and saturation \cite{wang2024lsle}, BREAD separates luminance enhancement, noise suppression, and chrominance recovery \cite{hu2023bread}, and DCC-Net combines grayscale content with colour histograms through pyramid colour embedding \cite{zhang2022dcc}. HVI-CIDNet processes intensity and chromatic information in the HVI colour space and exchanges features between branches \cite{yan2025hvi}. These methods improve enhancement through colour statistics, chrominance recovery, or colour-space design. Shallow features from pretrained convolutional networks also capture local boundaries and patterns \cite{zeiler2014visualizing}, and perceptual losses use pretrained features to guide image restoration \cite{johnson2016perceptual}. Colour representations and shallow texture features thus describe aspects of image content beyond luminance differences, with their contribution depending on how they enter subsequent image reconstruction.

CueDR-Net differs in how auxiliary representations are connected to output reconstruction. SMG guides appearance synthesis with predicted structure, DCC-Net combines grayscale content with colour statistics, and HVI-CIDNet exchanges intensity and chromatic features in a dedicated colour space. CueDR-Net retains an image stream and organizes local luminance variation, chroma magnitude, and shallow texture maps in a separate cue stream. Bidirectional interaction exchanges features throughout encoding and decoding. The resulting features support both the brightness adjustment path and the residual path, allowing the three cues to guide photometric adjustment and local content recovery throughout the network.

\subsection{Multi-Branch and Multi-Scale Image Restoration}

Multi-branch architectures organize restoration by processing role or signal representation. KinD assigns illumination adjustment and reflectance restoration to separate networks \cite{zhang2019kind}, SNR-Aware combines local and longer-range processing according to estimated signal reliability \cite{xu2022snr}, and HVI-CIDNet repeatedly exchanges intensity and chromatic features during encoding and decoding \cite{yan2025hvi}. In low-light polarimetric imaging, DStokes-CGCP designs separate branches around the characteristics of different Stokes components, illustrating how representation properties can inform processing roles \cite{duan2025dstokes}. These studies show that branch inputs, processing roles, and interaction points jointly shape reconstruction. For colour image enhancement, feature exchange between streams and reconstruction by output paths serve distinct roles that should be reflected in the overall network architecture.

Multi-scale processing provides neighbourhood information for structures of different spatial extents. MSRB uses parallel spatial filters to construct multi-scale responses \cite{li2018multi}, MIRNet exchanges information across feature streams at different resolutions \cite{zamir2020learning}, and coarse-to-fine methods progressively reconstruct content across image scales \cite{cho2021rethinking}. These approaches introduce scale diversity within blocks, across feature resolutions, or across processing stages. Feature fusion determines how the resulting responses contribute to restoration. SE and CBAM recalibrate channel or spatial responses \cite{hu2018squeeze,woo2018cbam}, while AFF and FFA-Net use attention to combine features \cite{dai2021attentional,qin2020ffa}. Multi-scale representations provide candidate information from different neighbourhoods, and feature weighting controls its contribution to reconstruction.

AMRRConv combines local transition responses and multi-scale processing within a single module. Compared with the parallel spatial filters in MSRB and the multi-resolution feature streams in MIRNet, AMRRConv applies factorized filters along the channel--height, channel--width, and height--width tensor planes at each scale. It combines these features with learnable response branches initialized with Sobel and Laplacian operators and aggregates candidate features through pixel wise fusion. The module is used at brightness adjustment map prediction and residual feature fusion. The former provides local transition and neighbourhood information for photometric adjustment, while the latter supports local-content reconstruction. Its distinction lies in integrating differential responses, multi-scale three-plane processing, and feature fusion at these two reconstruction~locations.

\section{Proposed Method}\label{sec:method}

\subsection{Overall Pipeline}

\begin{figure*}[pos=t!]
\centering
\includegraphics[width=0.98\textwidth]{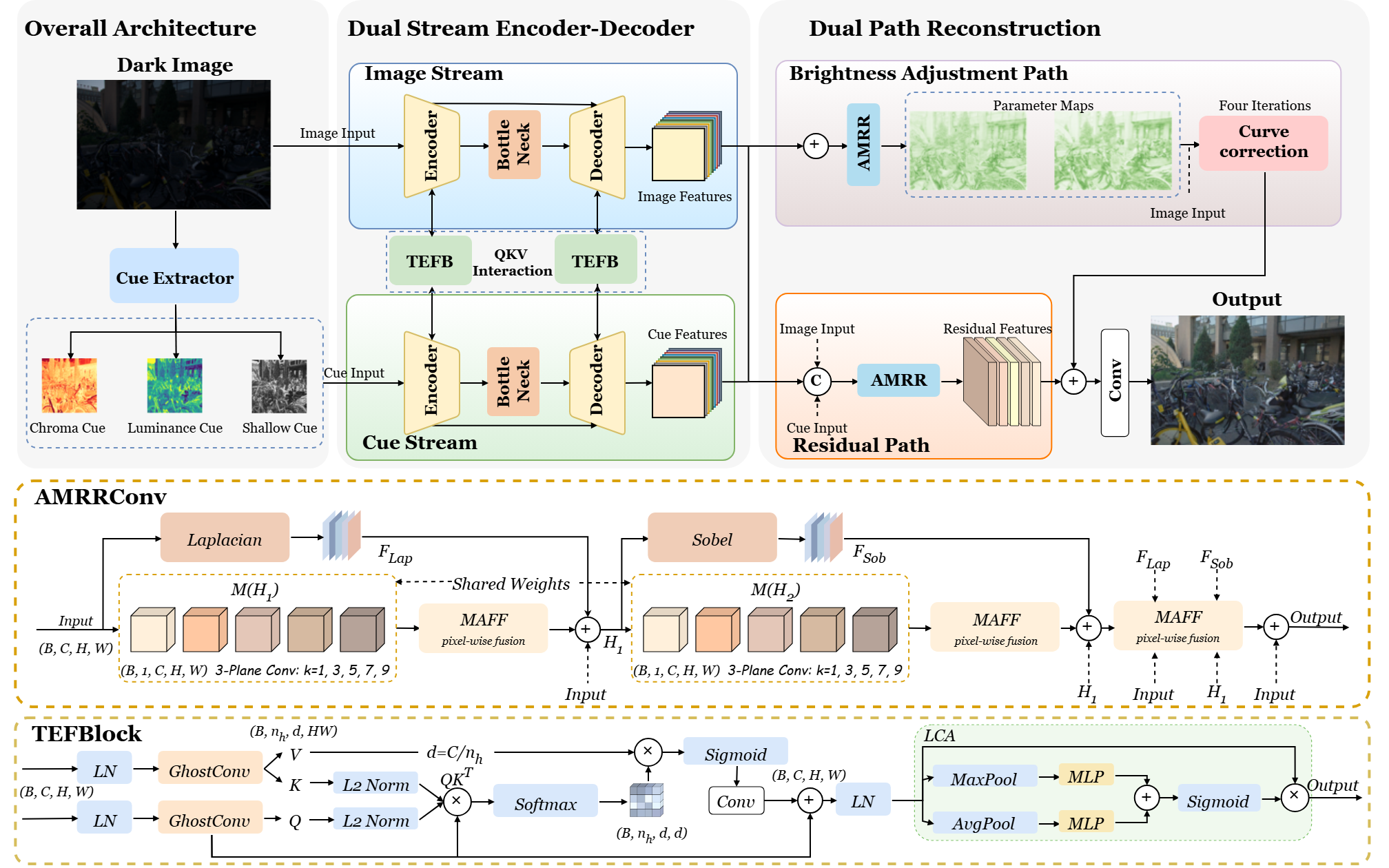}
\caption{Architecture of CueDR-Net and its AMRRConv and TEFBlock modules. The image and cue streams exchange features through TEFBlocks to support the brightness adjustment path and the residual path. AMRRConv refines features at brightness adjustment map prediction and residual feature fusion.}\label{fig1}
\end{figure*}

CueDR-Net uses two interacting feature streams to support two reconstruction paths, as shown in Fig.~\ref{fig1}. Given a low-light image $I_{in}\in[0,1]^{B\times3\times H\times W}$, a fixed extractor produces maps of local luminance variation, chroma magnitude, and shallow texture. The image and guidance maps enter separate image and cue streams. Lightweight Transformer Feature Extraction Blocks (TEFBlocks) exchange features between the streams during encoding and decoding, while skip connections preserve features from the corresponding encoder stages. The two streams organize image and guidance information, and the two paths use these features to construct the enhanced output. The brightness adjustment path predicts parameter maps for iterative pixel wise photometric adjustment. The residual path combines decoded features with shallow image and cue information to reconstruct local content. AMRRConv refines features at brightness adjustment map prediction and residual feature fusion. The outputs of the brightness adjustment path and the residual path are added and passed through a shared output mapping to produce the enhanced image.

\subsection{Cue-Guided Dual-Stream Interaction}

\begin{figure*}[pos=t]
\centering
\includegraphics[width=\textwidth]{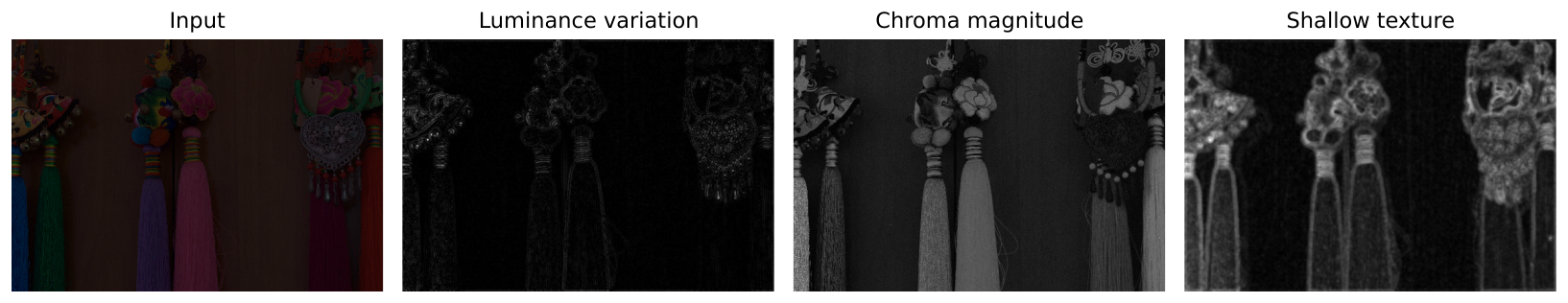}
\caption{Input and its three fixed guidance representations: local luminance variation, chroma magnitude, and shallow texture. Each cue is independently min--max normalized during extraction; grayscale intensity describes a response, not the original scene colour values.}\label{fig:ole_cues}
\end{figure*}

\textit{Fixed guidance cues}. The extractor represents the input through local luminance variation, chroma magnitude, and shallow texture, as illustrated in Fig.~\ref{fig:ole_cues}. Local luminance variation highlights intensity transitions, chroma magnitude measures departure from neutral colour, and shallow pretrained features capture local boundary and texture patterns. Each map is normalized independently within each image to emphasize relative spatial responses. The extractor remains fixed, while the subsequent cue stream learns how to use these representations. We use Lab to represent perceptual lightness $L^*$ separately from the opponent chromatic components $a$ and $b$ \cite{cie2018colorimetry}. This separation allows local brightness transitions and chroma magnitude to be extracted as distinct cues for brightness adjustment and colour preservation. For a single-channel map $U$, per-image min--max normalization is defined as
\begin{equation}
\mathcal{N}(U)(p)=
\frac{U(p)-\min_{q\in\Omega}U(q)}
{\max_{q\in\Omega}U(q)-\min_{q\in\Omega}U(q)+\epsilon},
\end{equation}
where $\Omega$ denotes the spatial image domain and $\epsilon=10^{-8}$ ensures numerical stability. Batch indices are omitted for clarity. The luminance variation cue uses a difference of Gaussians (DoG), whose centre-surround organization is related to receptive-field models in early vision \cite{enroth1966contrast}. Its spatial band-pass response suppresses slowly varying lightness and emphasizes local contrast transitions, providing structural guidance for enhancement. We obtain this cue from the absolute DoG response on $L^*$ and the chroma cue from the magnitude of $(a,b)$, giving
\begin{equation}
\begin{aligned}
\mathcal{F}_{s}&=\mathcal{N}\!\left(
\left|G_{5,1}*L^*-G_{9,2}*L^*\right|\right),\\
\mathcal{F}_{c}&=\mathcal{N}\!\left(\sqrt{a^2+b^2+\epsilon}\right),
\end{aligned}
\end{equation}
where $G_{5,1}$ and $G_{9,2}$ are Gaussian kernels normalized to sum to one, with sizes $5\times5$ and $9\times9$ and standard deviations of 1 and 2, respectively. Both convolutions use zero padding to preserve spatial dimensions.

The shallow texture cue is extracted using a frozen VGG-16 subnetwork pretrained on ImageNet. Before feature extraction, each RGB channel is instance-normalized within each image as
\begin{equation}
I_{in}^{\mathrm{IN},c}(p)=
\frac{I_{in}^{c}(p)-\mu_c}
{\sqrt{v_c+10^{-5}}},
\end{equation}
where $\mu_c$ and $v_c$ denote the spatial mean and population variance of channel $c$. The normalized image is passed to the frozen VGG-16 subnetwork defined by \texttt{features[:9]}. Its feature maps are averaged across channels, resized to the input resolution, and min--max normalized, giving
\begin{equation}
\mathcal{F}_{t}=\mathcal{N}\!\left(
\operatorname{Resize}_{H,W}\!\left(
\operatorname{Mean}_{ch}\!\left(
\phi_{\mathrm{VGG16}}^{[:9]}(I_{in}^{\mathrm{IN}})\right)\right)\right).
\end{equation}
Bilinear interpolation restores the input resolution with \texttt{align\_corners=False}. The three normalized maps are concatenated along the channel dimension to give the cue-stream input
\begin{equation}
\mathcal{F}_{aux}=\operatorname{Cat}(\mathcal{F}_{s},\mathcal{F}_{c},\mathcal{F}_{t})
\in\mathbb{R}^{B\times3\times H\times W}.
\end{equation}
Algorithm~\ref{alg:feature_generation} summarizes the cue extraction procedure.

\begin{algorithm}[t]
\caption{Auxiliary Feature Generation}\label{alg:feature_generation}
\KwIn{Low-light image $I_{in}\in[0,1]^{B\times3\times H\times W}$}
\KwOut{Auxiliary tensor $\mathcal{F}_{aux}\in\mathbb{R}^{B\times3\times H\times W}$}
\tcp{$\mathcal{N}$ normalizes each map independently per image.}
\tcp{Extract luminance variation and chroma magnitude.}
$(L^*,a,b)\gets\operatorname{RGB2LAB}(I_{in})$\;
$G_1\gets\mathcal{G}(5,1),\quad G_2\gets\mathcal{G}(9,2)$\;
$\mathcal{F}_{s}\gets\mathcal{N}(|G_1*L^*-G_2*L^*|)$\;
$\mathcal{F}_{c}\gets\mathcal{N}(\sqrt{a^2+b^2+10^{-8}})$\;
\tcp{Extract shallow texture with frozen VGG-16.}
$I^{\mathrm{IN}}_{in}\gets\operatorname{InstanceNorm}(I_{in};10^{-5})$\;
$\Phi_{vgg}\gets\phi_{\mathrm{VGG16}}^{[:9]}(I^{\mathrm{IN}}_{in})$\;
$\mathcal{F}_{t}\gets\mathcal{N}(\operatorname{Resize}_{H,W}(\operatorname{Mean}_{ch}(\Phi_{vgg})))$\;
\tcp{Concatenate normalized maps for the cue stream.}
$\mathcal{F}_{aux}\gets\operatorname{Cat}(\mathcal{F}_{s},\mathcal{F}_{c},\mathcal{F}_{t})$\;
\Return{$\mathcal{F}_{aux}$}
\end{algorithm}

\textit{Bidirectional feature exchange}. At each interaction stage, the image stream uses its own features as queries and obtains keys and values from the cue stream. The reverse update transfers image information to the cue stream. Each lightweight Transformer Feature Extraction Block (TEFBlock) models cross-stream channel relationships and then recalibrates the resulting features using lightweight channel attention (LCA). Let $X_I^l$ and $X_F^l$ denote the image- and cue-stream inputs. For the image-stream update, layer normalization operates across channels at each spatial position to give
\begin{equation}
\bar X_I^l=\operatorname{LN}_{pre}(X_I^l),\qquad
\bar X_F^l=\operatorname{LN}_{pre}(X_F^l).
\end{equation}
Ghost convolutions then project the normalized features to~form
\begin{equation}
Q=\mathcal{R}(g_q(\bar X_I^l)),\qquad
(K,V)=\mathcal{R}(g_{kv}(\bar X_F^l)).
\end{equation}
The reshape operator $\mathcal{R}$ arranges each query, key, and value tensor into $B\times n_h\times d\times HW$, where $n_h$ is the number of attention heads and $d=C/n_h$. Queries and keys are $\ell_2$-normalized over the $HW$ spatial positions before computing channel attention. The attention matrix and projected update are given by
\begin{equation}
\begin{aligned}
A&=\operatorname{Softmax}_{-1}\!\left(
\tau\,\operatorname{Norm}_{-1}(Q)
\operatorname{Norm}_{-1}(K)^T\right),\\
Z&=g_o\!\left(\mathcal{R}^{-1}\!\left(\sigma(AV)\right)\right),
\end{aligned}
\end{equation}
where $\tau$ is a learned temperature for each head, $\sigma$ denotes the sigmoid function, and $g_o$ is a $1\times1$ projection. Each head produces a $d\times d$ attention matrix, with softmax applied over key channels. The projected update $Z$ is added to the image-stream input, followed by layer normalization and LCA to~obtain
\begin{equation}
X_I^{l+1}=\operatorname{LCA}\!\left(
\operatorname{LN}_{post}(X_I^l+Z)\right).
\end{equation}
LCA recalibrates channel responses using spatially pooled statistics. For an input $U\in\mathbb{R}^{B\times C\times H\times W}$, spatial average pooling and max pooling produce the channel descriptors $u_{avg}$ and $u_{max}$. A shared MLP maps both descriptors to channel weights, giving
\begin{equation}
\begin{aligned}
m(z)&=W_2\operatorname{GELU}(W_1z),\\
\operatorname{LCA}(U)&=U\odot\sigma\!\left(m(u_{avg})+m(u_{max})\right).
\end{aligned}
\end{equation}
The shared MLP uses two bias-free $1\times1$ convolutions with channel widths $C\rightarrow\lfloor C/16\rfloor\rightarrow C$ and a GELU activation between them. The resulting channel weights are broadcast across spatial positions. The cue stream is updated by reversing the inputs to a separate TEFBlock with independent parameters. Both directions use the same features before either update, enabling simultaneous bidirectional exchange.

\subsection{Cue-Guided Dual-Path Reconstruction}

The image and cue features support two reconstruction paths with distinct output operations. The brightness adjustment path applies predicted three-channel adjustment maps to adjust input intensities through iterative pixel wise transformations. The residual path uses decoded features and shallow image and cue information to generate a correction from spatial context. AMRRConv refines features at brightness adjustment map prediction and residual feature fusion. The two path outputs are combined through a shared output mapping to form the enhanced image.

\textit{Brightness adjustment path}. Let $X_{up}^{3}$ and $Y_{up}^{3}$ denote the image and cue features at the final decoder resolution after skip addition and before the last TEFBlock pair. The AMRRConv module $\mathcal{M}_{\mathrm{curve}}$ processes their sum to predict four three-channel adjustment maps. Its output is clipped to $[-1,1]$ and split into the four maps as
\begin{equation}
\begin{aligned}
F_{\mathrm{param}}&=\mathcal{M}_{\mathrm{curve}}(X_{up}^{3}+Y_{up}^{3}),\\
(A_1,\ldots,A_4)&=\operatorname{Split}_{RGB}\!\left(
\operatorname{Clip}_{[-1,1]}(F_{\mathrm{param}})\right).
\end{aligned}
\end{equation}
Each parameter map $A_n\in[-1,1]^{B\times3\times H\times W}$ controls one iteration of the pixel wise transformation. Here, $K=4$ is a fixed implementation setting rather than an established optimum. Starting from $I^{(0)}=I_{in}$, we perform four iterations for $n=1,\ldots,4$~\cite{guo2020zero} according to
\begin{equation}
I^{(n)}=I^{(n-1)}+A_n\odot\left(
I^{(n-1)}-\left(I^{(n-1)}\right)^2\right).
\end{equation}
The output of the brightness adjustment path $I_{\mathrm{curve}}=I^{(4)}$ supplies spatially varying photometric adjustment.

\textit{Residual path and output fusion}. The final TEFBlock pair exchanges information between $X_{up}^{3}$ and $Y_{up}^{3}$ to produce decoded features $D_I=\operatorname{TEFBlock}_I(X_{up}^{3},Y_{up}^{3})$ and $D_F=\operatorname{TEFBlock}_F(Y_{up}^{3},X_{up}^{3})$. Separate reconstruction projections $P_I$ and $P_F$ map these features to three channels. Adding the input image and guidance maps through shallow skip connections gives
\begin{equation}
\begin{aligned}
X_{out}&=P_I(D_I)+I_{in},\\
Y_{out}&=P_F(D_F)+\mathcal{F}_{aux}.
\end{aligned}
\end{equation}
The projected outputs are concatenated with the original image and guidance maps, combining decoded content with shallow information. The AMRRConv module $\mathcal{M}_{\mathrm{res}}$ refines the combined features and predicts a three-channel correction given by
\begin{equation}
F_{\mathrm{res}}=\mathcal{M}_{\mathrm{res}}\!\left(
\operatorname{Cat}(X_{out},Y_{out},I_{in},\mathcal{F}_{aux})\right).
\end{equation}
The outputs of the two paths are added and passed through the shared output mapping $P_{\mathrm{out}}$ to give
\begin{equation}
I_{\mathrm{enh}}=P_{\mathrm{out}}(I_{\mathrm{curve}}+F_{\mathrm{res}}),
\end{equation}
where $P_{\mathrm{out}}$ consists of a $3\times3$ convolution from three to four channels, a sigmoid activation, and selection of the first three channels as the image output. Algorithm~\ref{alg:reconstruction} summarizes the dual-path reconstruction procedure.

\begin{algorithm}[t]
\caption{Cue-Guided \mbox{Dual-Path Reconstruction}}\label{alg:reconstruction}
\KwIn{Final upsampled, skip-connected features $X_{up}^{3},Y_{up}^{3}$; input $I_{in}$; auxiliary map $\mathcal{F}_{aux}$}
\KwOut{Enhanced image $I_{\mathrm{enh}}$}
\tcp{Predict four three-channel brightness adjustment maps.}
$F_{\mathrm{param}}\gets\mathcal{M}_{\mathrm{curve}}(X_{up}^{3}+Y_{up}^{3})$\;
$\{A_1,\ldots,A_4\}\gets\operatorname{Split}_{RGB}(\operatorname{Clip}_{[-1,1]}(F_{\mathrm{param}}))$\;
\tcp{Adjust input intensities through four iterations.}
$I^{(0)}\gets I_{in}$\;
\For{$n\gets1$ \KwTo $4$}{
  $I^{(n)}\gets I^{(n-1)}+A_n\odot\left(I^{(n-1)}-\left(I^{(n-1)}\right)^2\right)$\;
}
\tcp{Update both streams from the same pre-update features.}
$D_I\gets\operatorname{TEFBlock}_I(X_{up}^{3},Y_{up}^{3})$\;
$D_F\gets\operatorname{TEFBlock}_F(Y_{up}^{3},X_{up}^{3})$\;
\tcp{Combine decoded and shallow information to predict a correction.}
$X_{out}\gets P_I(D_I)+I_{in}$\;
$Y_{out}\gets P_F(D_F)+\mathcal{F}_{aux}$\;
$F_{\mathrm{res}}\gets\mathcal{M}_{\mathrm{res}}(\operatorname{Cat}(X_{out},Y_{out},I_{in},\mathcal{F}_{aux}))$\;
\tcp{Add the two path outputs and apply the shared output mapping.}
$I_{\mathrm{enh}}\gets P_{\mathrm{out}}(I^{(4)}+F_{\mathrm{res}})$\;
\Return{$I_{\mathrm{enh}}$}
\end{algorithm}

\subsection{Adaptive Multi-Scale Response Refinement Convolution}

AMRRConv combines local transition responses with multi-scale neighbourhood information to refine reconstruction features. Learnable differential branches capture local changes, factorized three-plane filters construct responses at different scales, and pixel wise fusion aggregates the resulting candidates. The module is applied at brightness adjustment map prediction and residual feature fusion. For an input $X\in\mathbb{R}^{B\times C\times H\times W}$, local filtering uses reshaped features of size $B\times1\times C\times H\times W$. The two AMRRConv module instances use input--output channel configurations of $(36,12)$ and $(12,3)$, respectively.

\textit{Learnable local transition responses}. Two branches use Laplacian and Sobel initializations to provide local transition responses. The Laplacian-initialized branch applies three independent, bias-free filters along the channel, height, and width axes, with initial kernels given by
\begin{equation}
K_{\mathrm{Lap},c}^{(0)}=[1,-2,1]^T,\qquad
K_{\mathrm{Lap},h}^{(0)}=K_{\mathrm{Lap},w}^{(0)}=[1,-2,1].
\end{equation}
The Sobel-initialized branch uses two spatial filters with support $1\times3\times3$ and a channel-axis filter with support $3\times1\times1$, with initial kernels given by
\begin{equation}
\begin{aligned}
K_{\mathrm{Sob},w}^{(0)}&=
\begin{bmatrix}1&0&-1\\2&0&-2\\1&0&-1\end{bmatrix},\quad
K_{\mathrm{Sob},h}^{(0)}=
\begin{bmatrix}1&2&1\\0&0&0\\-1&-2&-1\end{bmatrix},\\
K_{\mathrm{Sob},c}^{(0)}&=[1,0,-1]^T.
\end{aligned}
\end{equation}
All six filters are optimized jointly with the network. The absolute filtered responses are summed and passed through GELU to give
\begin{equation}
\mathcal{F}_{r}=\operatorname{GELU}\!\left(
\operatorname{Squeeze}\!\left(
\sum_{d\in\{c,h,w\}}|\hat X_r*\Theta_{r,d}|\right)\right),
\end{equation}
where $r\in\{\mathrm{Lap},\mathrm{Sob}\}$ identifies the response branch, $X_{\mathrm{Lap}}=X$, $X_{\mathrm{Sob}}=H_1$, and $\hat X_r=\operatorname{Unsqueeze}(X_r)$. The Laplacian branch processes the module input $X$, whereas the Sobel branch processes the first refinement output $H_1$. Taking absolute values retains response magnitudes rather than transition signs.

\textit{Multi-scale three-plane propagation}. Local interactions are organized along the channel and spatial axes of the feature tensor. Filtering on the channel--height and channel--width planes couples channel responses with spatial neighbours, while filtering on the height--width plane captures spatial variations. Applying these routes at multiple scales provides different neighbourhood support for weak boundaries and fine textures in low-light images. For each $k\in\{3,5,7,9\}$, a factorized branch aggregates responses on the channel--height, channel--width, and height--width tensor planes using filter supports of $k\times k\times1$, $k\times1\times k$, and $1\times k\times k$. Each branch also includes a pointwise filter, giving
\begin{equation}
\begin{aligned}
U_k={}&\operatorname{ReLU}(\mathcal{C}_{CH}^{k}(\hat X))
+\operatorname{ReLU}(\mathcal{C}_{CW}^{k}(\hat X))\\
&+\operatorname{ReLU}(\mathcal{C}_{HW}^{k}(\hat X))
+\operatorname{ReLU}(\mathcal{C}_{0,k}(\hat X)).
\end{aligned}
\end{equation}
The combined response is restored to four dimensions, added to the input, and processed by a grouped pointwise projection to give
\begin{equation}
M_k(X)=\operatorname{ReLU}\!\left(
\mathcal{D}_k(\operatorname{Squeeze}(U_k)+X)\right).
\end{equation}
The projection $\mathcal{D}_k$ maps $C$ channels to $2C$ using $C$ groups. A separate pointwise branch and the four propagation branches generate five candidate responses, defined as
\begin{equation}
\begin{aligned}
f_0(X)&=\operatorname{GELU}(\mathcal{D}_0(X)),\\
f_k(X)&=\operatorname{GELU}(M_k(X)),\\
\mathcal{M}(X)&=\{f_0(X),f_3(X),f_5(X),f_7(X),f_9(X)\},
\end{aligned}
\end{equation}
where $\mathcal{D}_0$ is a grouped pointwise projection from $C$ to $2C$ feature channels.

\textit{Pixel wise adaptive fusion}. Multi-scale adaptive feature fusion (MAFF) combines five candidates of equal shape using weights predicted at each spatial position. For the scale candidates, we relabel $\{f_0,f_3,f_5,f_7,f_9\}$ as $\{f_i\}_{i=1}^{5}$. A grouped projection first integrates their concatenation. Local convolutional context and spatially averaged context are then combined and modulated by channel and spatial attention as
\begin{equation}
\begin{aligned}
J&=\mathcal{P}_{int}([f_1,\ldots,f_5]),\\
T&=\operatorname{ReLU}(\mathcal{L}(J)+\mathcal{G}(J)),\\
\bar T&=T\odot\mathcal{A}_{ch}(T)\odot\mathcal{A}_{sp}(T).
\end{aligned}
\end{equation}
The integration projection $\mathcal{P}_{int}$ includes batch normalization and ReLU. Channel attention uses average pooling and pointwise projections, while spatial attention uses a $7\times7$ convolution. Both produce sigmoid gates. Five learned projections use the combined context to gate the candidates for weight prediction, giving
\begin{equation}
\begin{aligned}
\tilde f_i&=f_i\odot\sigma(\psi_i(\bar T)),\\
\ell&=\psi_w([\tilde f_1,\ldots,\tilde f_5]),\\
\omega_i(b,p)&=\frac{\exp(\ell_i(b,p))}
{\sum_{j=1}^{5}\exp(\ell_j(b,p))}.
\end{aligned}
\end{equation}
The logits have shape $B\times5\times H\times W$. Softmax across the five candidates produces weights that sum to one at each pixel and are shared across channels. These weights are applied to the original candidates, followed by a grouped pointwise output projection to give
\begin{equation}
\operatorname{MAFF}(f_1,\ldots,f_5)=
\mathcal{P}_{out}\!\left(\sum_{i=1}^{5}\omega_i\odot f_i\right),
\end{equation}
where $\mathcal{P}_{out}$ is a grouped $1\times1$ projection.

\textit{Staged refinement}. Starting from the module input $X$, the first round combines the fused multi-scale candidates with the Laplacian response and an input skip connection to produce $H_1$. The second round processes $H_1$ and incorporates its Sobel response together with skip connections from $X$ and $H_1$ to produce $H_2$. The candidate generators share parameters across the two rounds, while the MAFF modules use independent parameters. The two rounds and the module output are given by
\begin{equation}
\begin{aligned}
H_1&=\operatorname{MAFF}_1(\mathcal{M}(X))+\mathcal{F}_{\mathrm{Lap}}+X,\\
H_2&=\operatorname{MAFF}_2(\mathcal{M}(H_1))+\mathcal{F}_{\mathrm{Sob}}+X+H_1,\\
Y_{\mathrm{AMRRConv}}&=\operatorname{MAFF}_3
(H_2,H_1,\mathcal{F}_{\mathrm{Lap}},\mathcal{F}_{\mathrm{Sob}},X)\\
 &\quad+\operatorname{GELU}(\operatorname{Conv}_{res}(X)).
\end{aligned}
\end{equation}
The first two MAFF modules project the multi-scale candidates back to $C$ channels. The final MAFF aggregates $H_2$, $H_1$, the two response features, and the original input. Its output is added to a projected input connection. Both output projections map $C$ to $C_{out}$ using three groups.

\begin{figure*}[pos=tb]
\centering
\includegraphics[width=0.98\textwidth]{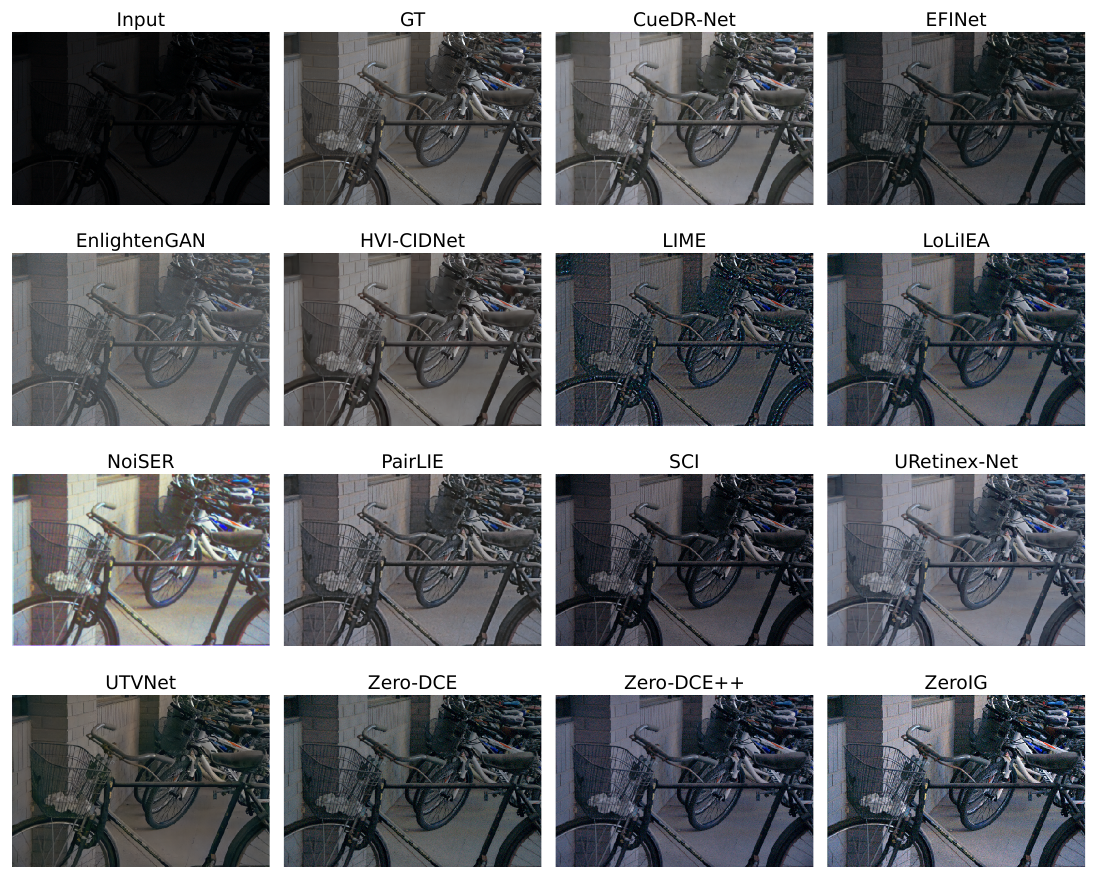}
\caption{Qualitative comparison of CueDR-Net with other state-of-the-art algorithms on the LOLv1 dataset.}\label{fig3}
\end{figure*}

\subsection{Loss Functions}

The training objective combines pixel fidelity, structural similarity, spatial regularity, and colour consistency. Let $E=I_{\mathrm{enh}}$ and $R=I_{gt}$ denote the enhanced and reference images. We use
\begin{equation}
\mathcal{L}_{\mathrm{total}}=
\mathcal{L}_1+\mathcal{L}_{\mathrm{ssim}}
+\mathcal{L}_{\mathrm{tv}}+\mathcal{L}_{\mathrm{hsv}},
\label{eq:loss_total}
\end{equation}
with unit weights for all four terms.

\textit{Pixel and structural fidelity}. The $L_1$ loss is the mean absolute difference between $E$ and $R$ over images, channels, and spatial positions, given by
\begin{equation}
\mathcal{L}_1=\frac{1}{BCHW}
\sum_{b,c,h,w}|E_{b,c,h,w}-R_{b,c,h,w}|.
\end{equation}

The structural term is
\begin{equation}
\mathcal{L}_{\mathrm{ssim}}=1-\operatorname{SSIM}(E,R),
\end{equation}
where SSIM is averaged over images, RGB channels, and valid window positions. SSIM uses an $11\times11$ Gaussian window with a standard deviation of 1.5 and constants $c_1=0.01^2$ and $c_2=0.03^2$ for images in $[0,1]$.

\textit{Spatial regularity}. The TV term penalizes squared differences between neighbouring pixels in the enhanced image along the horizontal and vertical directions. Each direction is normalized by its spatial difference count, and channel contributions are summed. The spatial differences are defined as
\begin{equation}
\begin{aligned}
\Delta_h E_{b,c,h,w}&=E_{b,c,h+1,w}-E_{b,c,h,w},\\
\Delta_w E_{b,c,h,w}&=E_{b,c,h,w+1}-E_{b,c,h,w}.
\end{aligned}
\end{equation}
The height-direction contribution to the TV term is
\begin{equation}
\mathcal{L}_{\mathrm{tv},h}=\frac{2}{B}
\frac{\sum_{b,c}\sum_{h=1}^{H-1}\sum_{w=1}^{W}(\Delta_h E_{b,c,h,w})^2}{(H-1)W},
\end{equation}
The width-direction contribution is
\begin{equation}
\mathcal{L}_{\mathrm{tv},w}=\frac{2}{B}
\frac{\sum_{b,c}\sum_{h=1}^{H}\sum_{w=1}^{W-1}(\Delta_w E_{b,c,h,w})^2}{H(W-1)},
\end{equation}
where $\mathcal{L}_{\mathrm{tv}}=\mathcal{L}_{\mathrm{tv},h}+\mathcal{L}_{\mathrm{tv},w}$ and $\sum_{b,c}$ sums over all $B$ images and $C=3$ channels.

\textit{Hue and saturation fidelity}. The colour term penalizes differences in hue and saturation after HSV conversion. Hue is expressed in radians, with the circular distance defined as
\begin{equation}
\begin{aligned}
\delta_H(b,p)&=|H_E(b,p)-H_R(b,p)|,\\
d_H(b,p)&=\min(\delta_H(b,p),2\pi-\delta_H(b,p)).
\end{aligned}
\end{equation}
The hue and saturation errors are combined as
\begin{equation}
\begin{aligned}
\mathcal{L}_{\mathrm{hsv}}=\frac{1}{BHW}\sum_{b,p}\Bigl[
&\lambda_{\mathrm{hue}}d_H(b,p)\\
&+\lambda_{\mathrm{sat}}|S_E(b,p)-S_R(b,p)|\Bigr].
\end{aligned}
\end{equation}
The hue and saturation terms have equal weights, with $\lambda_{\mathrm{hue}}=\lambda_{\mathrm{sat}}=1$.

\begin{figure*}[pos=tb]
\centering
\includegraphics[width=0.98\textwidth]{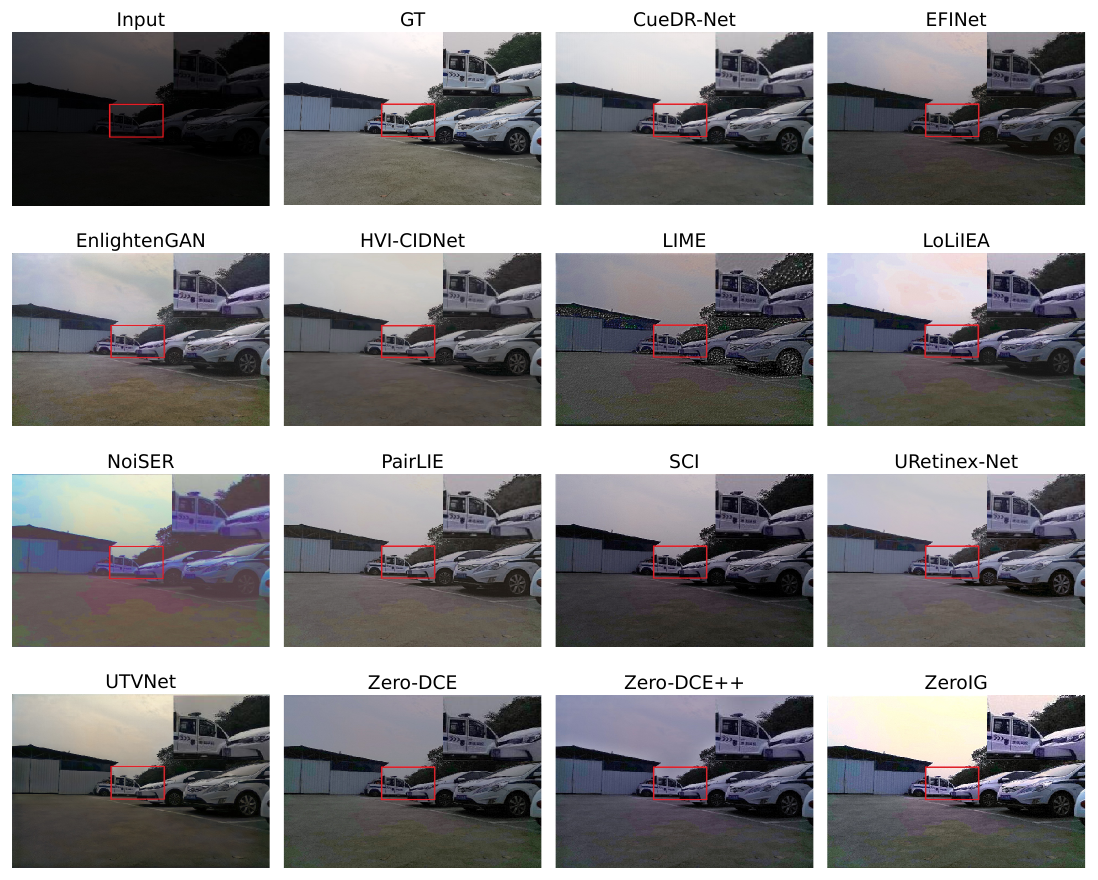}
\caption{Qualitative comparison of CueDR-Net with other state-of-the-art algorithms on the LSRW-HUAWEI dataset. The red rectangle marks the same local region in each output.}\label{fig4}
\end{figure*}

\begin{table*}[pos=tbp]
\caption{Quantitative comparison of CueDR-Net with other state-of-the-art algorithms on the LOLv1 dataset. Deep orange and light orange mark the best and second-best results in each column, respectively.}\label{tbl1}
{\setlength\tabcolsep{3pt}%
\resizebox{\textwidth}{!}{%
\begin{tabular}{@{}l*{7}{>{\centering\arraybackslash}p{1.9cm}}@{}}
\toprule
Method & PSNR$\uparrow$ & SSIM$\uparrow$ & LOE$\downarrow$ & DE$\uparrow$ & EME$\uparrow$ & Params (M)$\downarrow$ & FLOPs (G)$\downarrow$ \\
\midrule
ZeroDceP \cite{li2021learning} & 15.3809 & 0.6100 & 25.9252 & 1.9906 & 19.5745 & 0.011 & \best{0.05} \\
ZeroDCE \cite{guo2020zero} & 14.9857 & 0.6009 & 23.7979 & 1.8426 & 19.5491 & 0.079 & 19.01 \\
SCI \cite{ma2022toward} & 13.8701 & 0.5628 & \seco{6.8127} & 1.9228 & 20.1908 & \best{0.0003} & \seco{0.13} \\
RUAS \cite{liu2021retinex} & 11.3339 & 0.4577 & \best{0.3253} & 1.4712 & \seco{20.8448} & 0.003 & 0.78 \\
EnlightenGAN \cite{jiang2021enlightengan} & 17.0475 & 0.7162 & 64.7671 & 1.6520 & 2.9972 & 8.637 & 61.03 \\
FmrNet \cite{chen2024fmr} & 18.8707 & 0.7922 & 27.3718 & 2.2879 & 2.9731 & 0.197 & 376.36 \\
LIME \cite{guo2016lime} & 14.0122 & 0.5013 & 102.1583 & 1.8400 & \best{22.2654} & -- & -- \\
FRRNET \cite{chen2024frr} & 18.2786 & 0.7500 & 24.2262 & 1.9453 & 3.8288 & 0.018 & 0.95 \\
UTVNet \cite{zheng2021adaptive} & 15.6201 & 0.7354 & 30.5412 & 2.0346 & 4.5721 & 7.751 & 77.93 \\
ChebyLighter \cite{pan2022chebylighter} & 19.8546 & 0.7791 & 18.2176 & 2.1641 & 3.5892 & 0.178 & 42.00 \\
EFINet \cite{liu2022efinet} & 13.4802 & 0.6414 & 25.1893 & 1.6471 & 5.3525 & 0.129 & 34.35 \\
PairLIE \cite{fu2023learning} & 18.5856 & 0.7652 & 51.8890 & 1.9091 & 3.7587 & 0.342 & 81.84 \\
LoLiIEA \cite{perez2023loli} & 17.1161 & 0.5855 & 38.0141 & 2.1105 & 6.9952 & 26.593 & 105.76 \\
NoiSER \cite{zhang2024noise} & 17.3610 & 0.7012 & 52.1913 & 2.3581 & 2.5638 & \seco{0.002} & 0.43 \\
URetinex\_Net \cite{wu2022uretinex} & 20.4945 & 0.8372 & 29.9825 & 2.1280 & 3.4536 & 0.340 & 144.92 \\
ZeroIG \cite{shi2024zero} & 17.9818 & 0.5044 & 14.4277 & \best{2.5593} & 20.5418 & 0.087 & 118.73 \\
HVI-CIDNet \cite{yan2025hvi} & \seco{23.8082} & \best{0.8565} & 42.3837 & 2.3599 & 4.1153 & 1.976 & 29.78 \\
\textbf{CueDR-Net (Ours)} & \best{24.4921} & \seco{0.8522} & 29.6277 & 2.3495 & 3.3851 & 6.360 & 103.64 \\
\bottomrule
\end{tabular}}}
\end{table*}

\section{Experiments}
We evaluate CueDR-Net on LOLv1 \cite{wei2018deep}, LSRW-HUAWEI, and LSRW-NIKON \cite{hai2023r2rnet} to assess enhancement across scenes and camera sources. LOLv1 provides 485 paired low-light and normal-light images for training and 15 pairs for testing, covering indoor and outdoor scenes. LSRW-HUAWEI and LSRW-NIKON contain paired images captured with a HUAWEI P40 Pro and a NIKON D7500, respectively. These datasets support comparisons of brightness recovery, colour appearance, and local detail under different capture conditions.

CueDR-Net is implemented in PyTorch 2.1.0 with CUDA 12.1 and trained on two NVIDIA RTX 4090 GPUs with 24 GB of memory each. Training follows two stages using the same objective in Eq.~\eqref{eq:loss_total}. In the first stage, Adam uses a batch size of 6 and an initial learning rate of $5\times10^{-3}$, which is halved every 50 epochs over 350 epochs. In the second stage, training continues for a further 1,000 epochs with cosine annealing from $1\times10^{-4}$ to $1\times10^{-6}$. Training images are randomly cropped to $192\times192$ pixels and scaled to $[0,1]$.

\subsection{Comparative Results}

\begin{figure*}[pos=tb]
\centering
\includegraphics[width=0.98\textwidth]{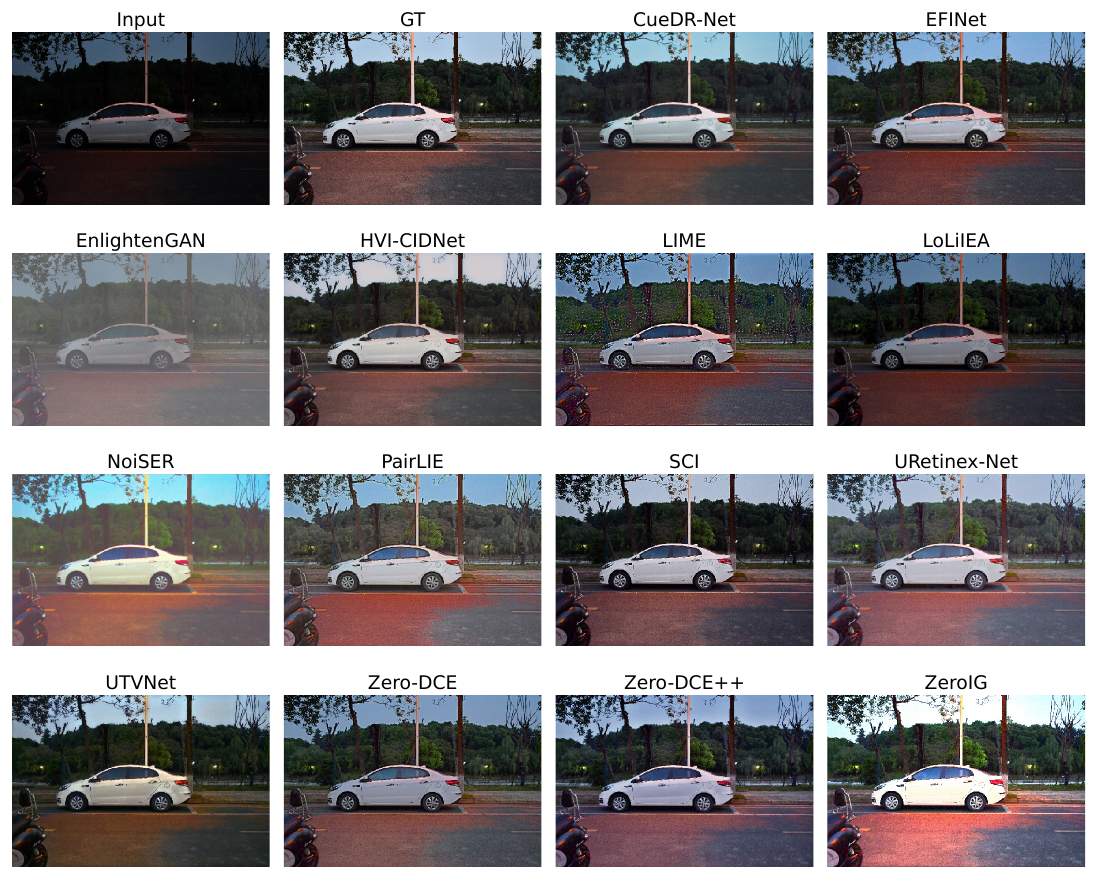}
\caption{Qualitative comparison of CueDR-Net with other state-of-the-art algorithms on the LSRW-N dataset.}\label{fig6}
\end{figure*}

\begin{table*}[pos=tbp]
\caption{Quantitative comparison of CueDR-Net with other state-of-the-art algorithms on the LSRW-H dataset. Deep orange and light orange mark the best and second-best results in each column, respectively.}\label{tbl2}
{\setlength\tabcolsep{3pt}%
\resizebox{\textwidth}{!}{%
\begin{tabular}{@{}l*{7}{>{\centering\arraybackslash}p{1.9cm}}@{}}
\toprule
Method & PSNR$\uparrow$ & SSIM$\uparrow$ & LOE$\downarrow$ & DE$\uparrow$ & EME$\uparrow$ & Params (M)$\downarrow$ & FLOPs (G)$\downarrow$ \\
\midrule
ZeroDceP \cite{li2021learning} & 16.5502 & 0.4744 & 25.6661 & 1.9038 & 17.2490 & 0.011 & \best{0.05} \\
ZeroDCE \cite{guo2020zero} & 16.3458 & 0.4655 & 23.7504 & 1.8220 & 17.1155 & 0.079 & 19.01 \\
SCI \cite{ma2022toward} & 15.0965 & 0.4190 & \seco{7.2390} & 1.8007 & 17.8695 & \best{0.0003} & \seco{0.13} \\
RUAS \cite{liu2021retinex} & 12.5352 & 0.3450 & \best{1.1905} & 1.2653 & \seco{18.2901} & 0.003 & 0.78 \\
EnlightenGAN \cite{jiang2021enlightengan} & 18.0427 & 0.5307 & 61.3522 & 1.7745 & 2.4203 & 8.637 & 61.03 \\
FmrNet \cite{chen2024fmr} & 19.0461 & \seco{0.5779} & 24.2036 & 2.2503 & 2.5166 & 0.197 & 376.36 \\
LIME \cite{guo2016lime} & 14.9498 & 0.3462 & 115.0459 & 1.8318 & \best{20.7659} & -- & -- \\
FRRNET \cite{chen2024frr} & 19.0108 & 0.5309 & 23.4471 & 2.0170 & 3.4681 & 0.018 & 0.95 \\
UTVNet \cite{zheng2021adaptive} & 18.2295 & 0.5552 & 23.7439 & 2.2151 & 4.2250 & 7.751 & 77.93 \\
ChebyLighter \cite{pan2022chebylighter} & 17.8063 & 0.5342 & 52.4886 & 2.1960 & 3.6836 & 0.178 & 42.00 \\
EFINet \cite{liu2022efinet} & 14.8979 & 0.4918 & 22.2018 & 1.6330 & 4.5798 & 0.129 & 34.35 \\
PairLIE \cite{fu2023learning} & 19.0102 & 0.5587 & 40.4256 & 1.8325 & 2.8535 & 0.342 & 81.84 \\
LoLiIEA \cite{perez2023loli} & 15.9741 & 0.4932 & 23.2257 & 1.6996 & 8.0802 & 26.593 & 105.76 \\
NoiSER \cite{zhang2024noise} & 15.7901 & 0.5299 & 38.6788 & 1.9330 & 1.5256 & \seco{0.002} & 0.43 \\
URetinex\_Net \cite{wu2022uretinex} & \seco{19.5233} & 0.5502 & 29.0030 & 2.1873 & 4.5329 & 0.340 & 144.92 \\
ZeroIG \cite{shi2024zero} & 16.8910 & 0.4163 & 10.4842 & 2.2625 & 17.9813 & 0.087 & 118.73 \\
HVI-CIDNet \cite{yan2025hvi} & 18.8996 & 0.5649 & 30.2100 & \best{2.3922} & 4.3475 & 1.976 & 29.78 \\
\textbf{CueDR-Net (Ours)} & \best{20.6867} & \best{0.6126} & 34.8694 & \seco{2.3109} & 2.5426 & 6.360 & 103.64 \\
\bottomrule
\end{tabular}}}
\end{table*}

\subsubsection{Qualitative Results}
We compare representative methods on LOLv1, LSRW-HUAWEI, and LSRW-NIKON in Figs.~\ref{fig3}, \ref{fig4}, and \ref{fig6} respectively. The selected scenes are used to examine the recovery of dark-region visibility, the stability of colour appearance, and the preservation of local contours and textures.

\textit{LOLv1}. The bicycle scene in Fig.~\ref{fig3} combines deep shadows with fine spokes and object boundaries. EFINet, LIME, SCI, UTVNet, and Zero-DCE leave parts of the shadows under-enhanced. EnlightenGAN and NoiSER increase brightness but show reduced spoke visibility or colour imbalance. LoLiIEA, URetinex-Net, ZeroIG, and Zero-DCE++ exhibit colour deviations, while PairLIE produces a brighter image with less distinct thin contours. CueDR-Net improves shadow visibility while preserving the fine bicycle spokes and maintaining a balanced overall colour appearance.

\textit{LSRW-HUAWEI}. The vehicle-yard scene in Fig.~\ref{fig4} includes smooth sky, shaded vehicle surfaces, and small cabin details. LIME and NoiSER show spatial artifacts and uneven colour. SCI and Zero-DCE leave shaded areas relatively dark, while EnlightenGAN, LoLiIEA, and Zero-DCE++ introduce colour casts on neutral vehicle surfaces. HVI-CIDNet and PairLIE retain more neutral colours but differ in shadow brightness. CueDR-Net improves visibility along vehicle contours and preserves local cabin details, although sky and shadow brightness still differ from the reference.

\textit{LSRW-NIKON}. Fig.~\ref{fig6} shows a roadside vehicle with dark trees and a broad road surface. SCI and LoLiIEA retain dark background regions, whereas ZeroIG strongly brightens the vehicle and pavement. LIME introduces spatial artifacts, and NoiSER shows a pronounced warm cast. EnlightenGAN appears lower in contrast; HVI-CIDNet and PairLIE provide more restrained enhancement. CueDR-Net preserves the vehicle outline and visible tree transitions while improving the shaded content, although a cooler chromatic appearance remains relative to the reference. These examples expose exposure and colour trade-offs rather than establishing a uniform qualitative ranking.

\begin{table*}[pos=tbp]
\caption{Quantitative comparison of CueDR-Net with other state-of-the-art algorithms on the LSRW-N dataset. Deep orange and light orange mark the best and second-best results in each column, respectively.}\label{tbl3}
{\setlength\tabcolsep{3pt}%
\resizebox{\textwidth}{!}{%
\begin{tabular}{@{}l*{7}{>{\centering\arraybackslash}p{1.9cm}}@{}}
\toprule
Method & PSNR$\uparrow$ & SSIM$\uparrow$ & LOE$\downarrow$ & DE$\uparrow$ & EME$\uparrow$ & Params (M)$\downarrow$ & FLOPs (G)$\downarrow$ \\
\midrule
ZeroDceP \cite{li2021learning} & 15.6097 & 0.4294 & 46.2506 & \seco{1.2562} & 9.2870 & 0.011 & \best{0.05} \\
ZeroDCE \cite{guo2020zero} & 15.0306 & 0.4145 & 51.6386 & 1.0757 & 9.3504 & 0.079 & 19.01 \\
SCI \cite{ma2022toward} & 15.2775 & 0.3852 & 25.9748 & 1.2082 & 11.2255 & \best{0.0003} & \seco{0.13} \\
RUAS \cite{liu2021retinex} & 13.7949 & 0.3737 & \best{1.9827} & 0.6629 & \best{13.1106} & 0.003 & 0.78 \\
EnlightenGAN \cite{jiang2021enlightengan} & 15.3151 & 0.4332 & 91.7290 & 0.7167 & 3.0026 & 8.637 & 61.03 \\
FmrNet \cite{chen2024fmr} & 16.5829 & \seco{0.4783} & 24.3769 & 0.9551 & 4.0877 & 0.197 & 376.36 \\
LIME \cite{guo2016lime} & 14.4994 & 0.3270 & 89.2000 & \best{1.2573} & 12.1026 & -- & -- \\
FRRNET \cite{chen2024frr} & 16.2898 & 0.4614 & 26.0151 & 1.0270 & 3.5325 & 0.018 & 0.95 \\
UTVNet \cite{zheng2021adaptive} & 13.6543 & 0.4277 & 28.2466 & 1.0129 & 6.8350 & 7.751 & 77.93 \\
ChebyLighter \cite{pan2022chebylighter} & 12.9789 & 0.3905 & 87.8313 & 1.2482 & 4.8674 & 0.178 & 42.00 \\
EFINet \cite{liu2022efinet} & 15.7816 & 0.4258 & 36.8270 & 1.0564 & 5.8542 & 0.129 & 34.35 \\
PairLIE \cite{fu2023learning} & 15.5299 & 0.4309 & 60.0751 & 1.0621 & 4.4785 & 0.342 & 81.84 \\
LoLiIEA \cite{perez2023loli} & 13.8871 & 0.4139 & 25.5192 & 0.9722 & \seco{12.7845} & 26.593 & 105.76 \\
NoiSER \cite{zhang2024noise} & 15.5844 & 0.4450 & 41.8533 & 0.4797 & 1.9870 & \seco{0.002} & 0.43 \\
URetinex\_Net \cite{wu2022uretinex} & \seco{16.6411} & 0.4506 & 26.3874 & 0.9570 & 4.5285 & 0.340 & 144.92 \\
ZeroIG \cite{shi2024zero} & 12.6374 & 0.3435 & 34.9467 & 0.9107 & 10.3915 & 0.087 & 118.73 \\
HVI-CIDNet \cite{yan2025hvi} & 15.4478 & 0.4529 & \seco{20.6919} & 1.1225 & 5.1495 & 1.976 & 29.78 \\
\textbf{CueDR-Net (Ours)} & \best{16.7462} & \best{0.4799} & 24.2339 & 1.0408 & 4.1418 & 6.360 & 103.64 \\
\bottomrule
\end{tabular}}}
\end{table*}

\subsubsection{Quantitative Results}

Tables~\ref{tbl1}--\ref{tbl3} report full-reference and no-reference results on LOLv1, LSRW-HUAWEI, and LSRW-NIKON. PSNR measures pixel-level reconstruction accuracy. SSIM assesses structural similarity. LOE, DE, and EME provide additional no-reference assessments of illumination ordering, information content, and enhancement strength. These metrics are interpreted together with the cue, dual-path, and AMRRConv ablations to evaluate the complete reconstruction system.

Tables~\ref{tbl1}--\ref{tbl3} show that CueDR-Net achieves the highest PSNR on all three benchmarks. On LOLv1, it obtains 24.4921\,dB, exceeding HVI-CIDNet by 0.6839\,dB, with the second-best SSIM of 0.8522. On LSRW-HUAWEI and LSRW-NIKON, it achieves the highest PSNR/SSIM of 20.6867\,dB/0.6126 and 16.7462\,dB/0.4799, respectively. Its PSNR gains over URetinex-Net are 1.1634\,dB and 0.1051\,dB on these two datasets. These results demonstrate consistent pixel fidelity and competitive structural preservation across different capture conditions. CueDR-Net uses 6.360M parameters and 103.64 GFLOPs, requiring more resources than lightweight methods such as HVI-CIDNet. This cost reflects the processing of two interacting feature streams and two reconstruction paths. The leading PSNR results and strong SSIM scores, together with the improved shadow visibility and preserved local detail in Figs.~\ref{fig3}--\ref{fig6}, support this emphasis on enhancement quality.

\begin{figure*}[pos=tb]
\centering
\includegraphics[width=0.98\textwidth]{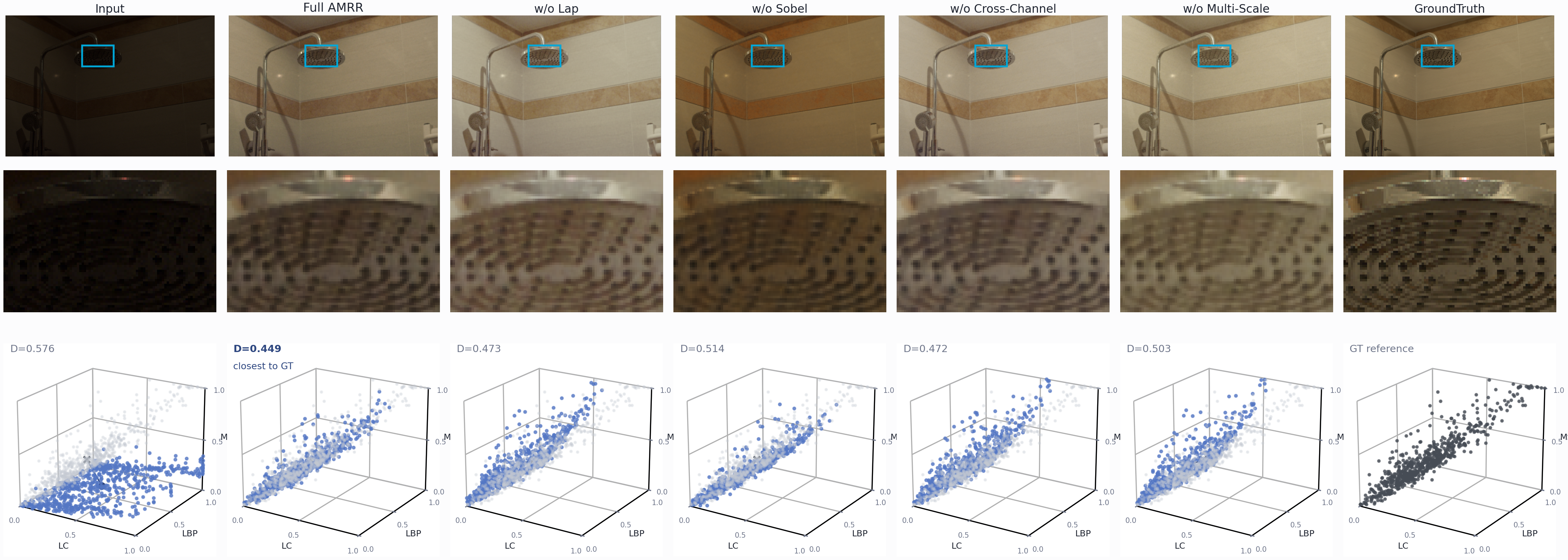}
\caption{AMRRConv component ablation in local contrast (LC), local binary pattern (LBP), and multi-scale Gaussian detail (MSGD) space. Blue: result descriptors; light gray: GT. $D_{\mathrm{feat}}$ is the mean $L_1$ deviation between corresponding normalized pixel descriptors; lower is better.}\label{fig:component_structure}
\end{figure*}

\begin{figure*}[pos=tb]
\centering
\includegraphics[width=0.94\textwidth]{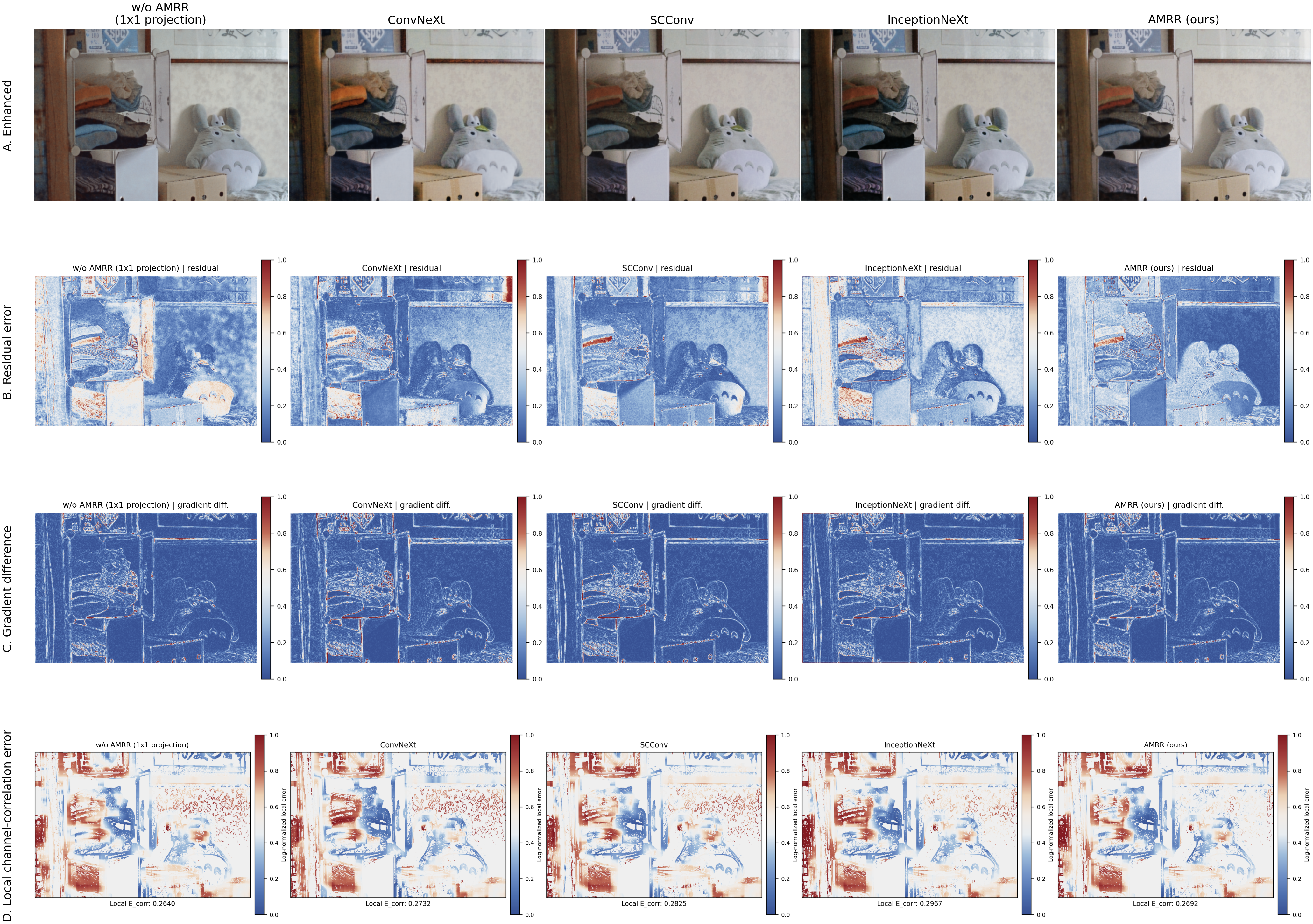}
\caption{AMRRConv module replacement (Table~\ref{tbl:module_effect}). Rows: \textbf{A}, enhanced images; \textbf{B}, residual error $R=|I_{\mathrm{enhanced}}-I_{\mathrm{GT}}|$; \textbf{C}, gradient difference $D_g=|G(I_{\mathrm{enhanced}})-G(I_{\mathrm{GT}})|$ with $G(I)=|\nabla I|$; \textbf{D}, local channel-correlation error $E_{\mathrm{corr}}^{\mathrm{local}}(p)=\sum_{ij\in\{RG,RB,GB\}}|\rho_{\mathrm{enh},w_p}^{ij}-\rho_{\mathrm{GT},w_p}^{ij}|$, where $\rho^{ij}_{w_p}$ is the Pearson correlation of channel-$i$ and channel-$j$ Sobel-gradient magnitudes within $w_p$. Row~\textbf{D} uses log normalization over strong GT-structure regions; warmer colours indicate larger error.}\label{fig:module_visual}
\end{figure*}

\subsection{Ablation Study}\label{sec:ablation}
We train the propagation, guidance-cue, and reconstruction-path variants on LOLv1 for 250 epochs using Adam with a batch size of 6. The initial learning rate is $5\times10^{-3}$ and is halved every 50 epochs. Training uses paired random crops of $192\times192$ pixels and matched horizontal and vertical flips. All variants in these groups use the objective in Eq.~\eqref{eq:loss_total}. To compare multiple design variants at a manageable computational cost, we use a common schedule of 250 epochs rather than the full two-stage procedure adopted for the main benchmark comparisons.

The LOLv1 ablations examine the training objective, AMRRConv, guidance cues, and reconstruction paths. For the propagation, module, cue, and path comparisons in Tables~\ref{tbl:propagation}--\ref{tbl:dual_path}, we report GMSD~\cite{xue2014gmsd}, CIEDE2000~\cite{sharma2005ciede}, and luminance MAE alongside PSNR and SSIM to assess gradient similarity, colour error, and brightness error. With $Y=0.299R+0.587G+0.114B$, luminance MAE is defined as $\frac{1}{HW}\sum_p|Y_{\mathrm{enh}}(p)-Y_{\mathrm{GT}}(p)|$. Scores are computed from float RGB outputs using a common $592\times400$ crop and averaged over the 15 LOLv1 test images. Lower values are better for the three additional metrics.

\begin{figure*}[pos=tb]
\centering
\includegraphics[width=0.96\textwidth]{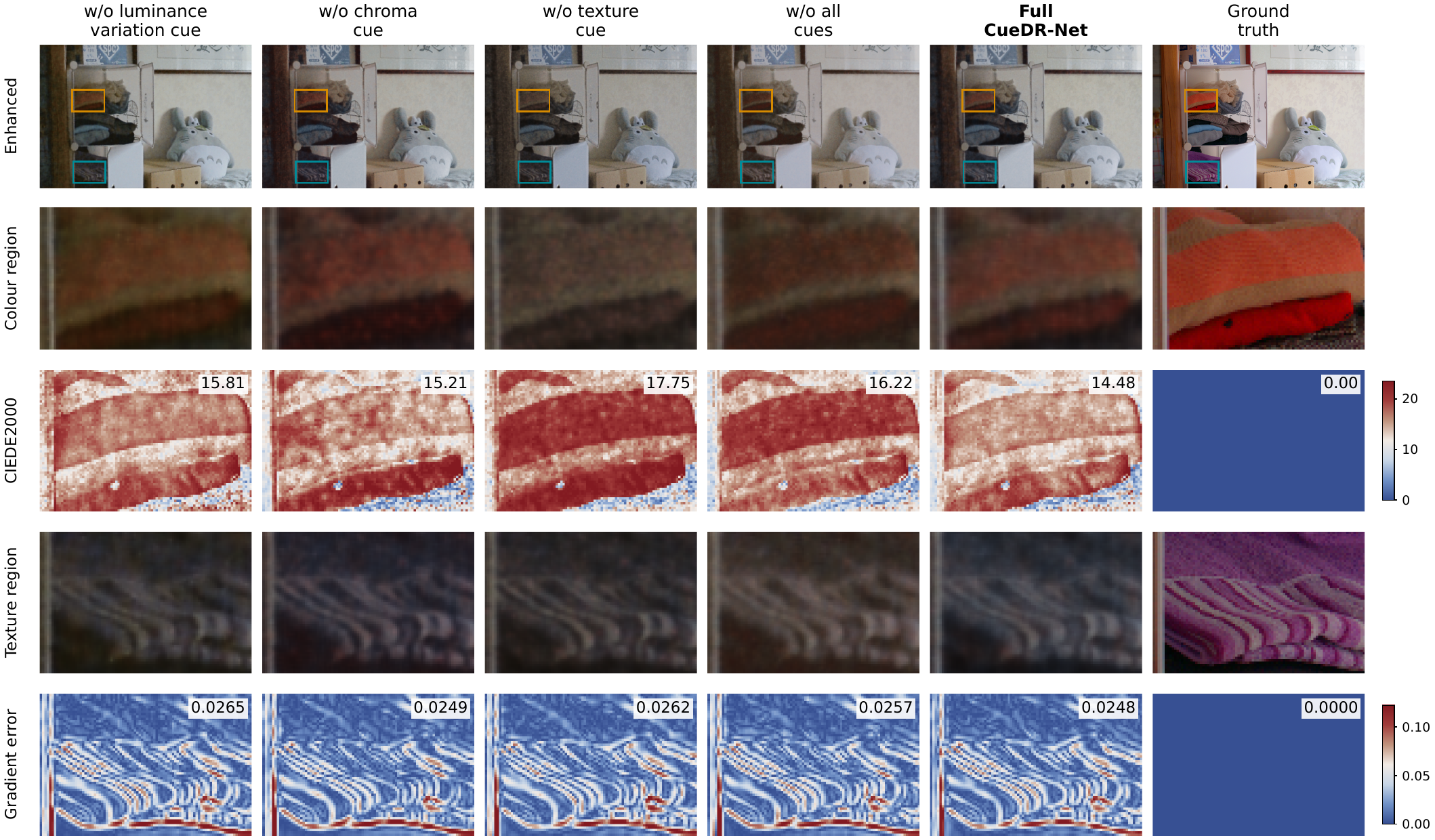}
\caption{Cue removal in common colour (orange) and texture (cyan) regions. Rows: outputs, colour crops, CIEDE2000 maps, texture crops, and gradient-error maps. Each error measure uses a shared linear scale; labels give mean patch errors.}\label{fig:ole_guidance}
\end{figure*}

\subsubsection{Training Objective}
Table~\ref{tbl4} compares the loss configurations on LOLv1. The $L_1$+SSIM+TV+HSV objective achieves the best PSNR and SSIM of 21.84\,dB and 0.8080. With the other terms unchanged, removing TV reduces these scores to 21.46\,dB and 0.8052, while removing HSV lowers them to 19.94\,dB and 0.7829. These comparisons support retaining both spatial regularization and hue--saturation supervision. Adding the exploratory exposure term (EXP) to the $L_1$+SSIM+TV baseline reduces both PSNR and SSIM, so it is excluded from the final objective.
\begin{table}[pos=!htbp]
\centering
\caption{Ablation study on the loss function configuration for CueDR-Net (LOLv1 dataset). EXP is evaluated only as an exploratory term and is not used by the final model. Deep orange and light orange mark the best and second-best quantitative results in each column, respectively.}\label{tbl4}
\begin{tabular*}{\columnwidth}{@{} LLLLLLL @{} }
\toprule
$L_1$ & SSIM & TV & EXP & HSV & PSNR$\uparrow$ & SSIM$\uparrow$ \\
\midrule
$\surd$ & $\surd$ & $\surd$ & $\surd$ & $\times$ & 17.66 & 0.7583 \\
$\times$ & $\surd$ & $\surd$ & $\surd$ & $\times$ & 15.95 & 0.7345 \\
$\surd$ & $\surd$ & $\surd$ & $\times$ & $\times$ & 19.94 & 0.7829 \\
$\surd$ & $\surd$ & $\times$ & $\surd$ & $\times$ & 17.57 & 0.7453 \\
$\surd$ & $\surd$ & $\times$ & $\times$ & $\surd$ & \seco{21.46} & \seco{0.8052} \\
$\surd$ & $\surd$ & $\surd$ & $\times$ & $\surd$ & \best{21.84} & \best{0.8080} \\
$\surd$ & $\times$ & $\surd$ & $\surd$ & $\times$ & 18.56 & 0.7293 \\
\bottomrule
\end{tabular*}
\end{table}

\begin{figure*}[pos=tb]
\centering
\includegraphics[width=0.96\textwidth]{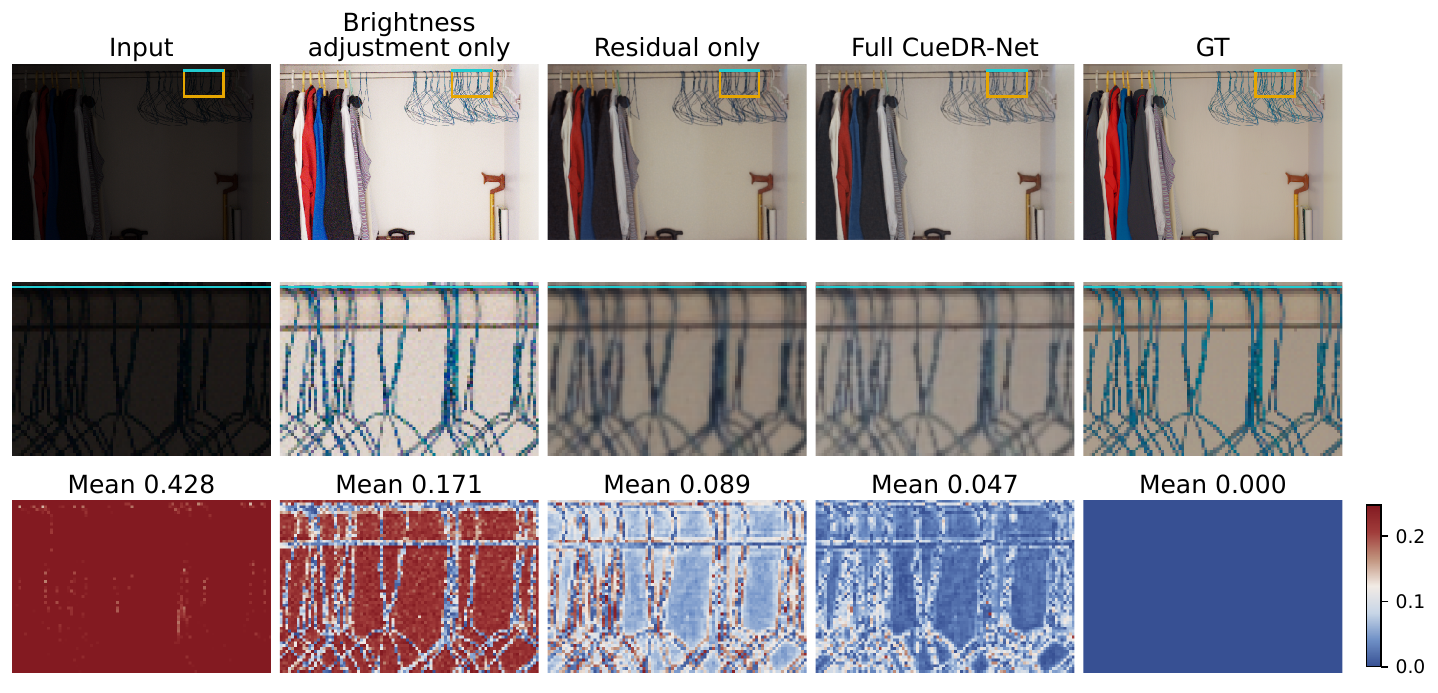}
\caption{Single-path and full reconstruction. Rows: outputs, common crops, and absolute luminance error on a shared linear scale. The cyan line marks the scanline in Fig.~\ref{fig:ole_profile}.}\label{fig:ole_paths}
\vspace{8pt}
\includegraphics[width=0.94\textwidth]{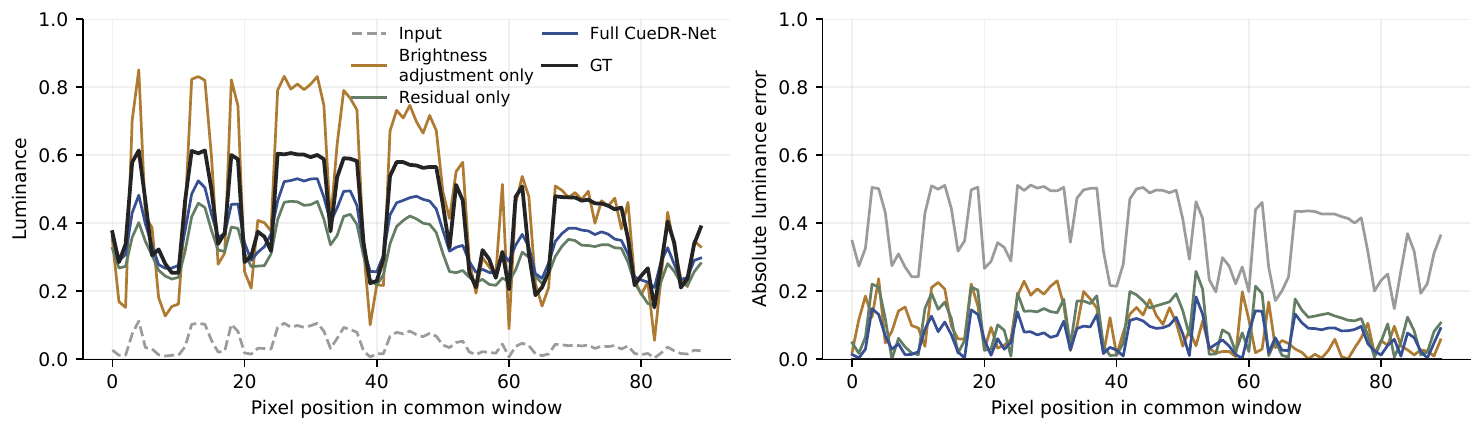}
\caption{Luminance and absolute luminance error along the cyan scanline in Fig.~\ref{fig:ole_paths}, chosen from reference transitions. All outputs use identical pixel positions without alignment.}\label{fig:ole_profile}
\end{figure*}

\begin{figure*}[pos=tb]
\centering
\includegraphics[width=\textwidth]{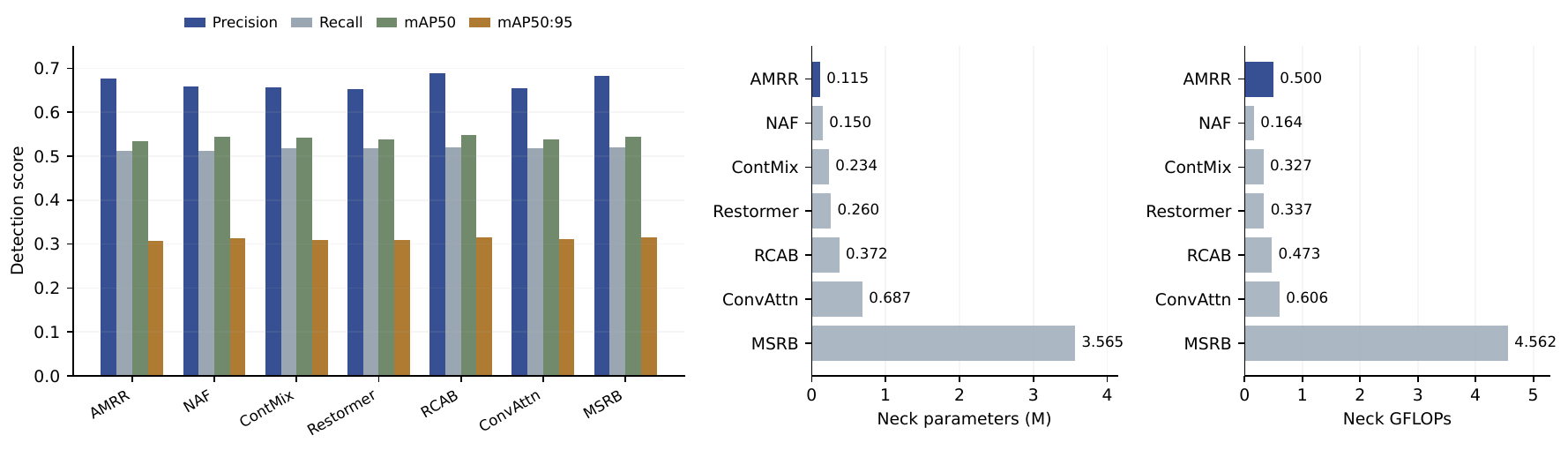}
\caption{AMRRConv and alternative modules at two YOLOv11 neck locations on ExDark. Left: detection performance. Middle: neck parameters. Right: operation estimates.}\label{fig:yolo}
\end{figure*}

\subsubsection{AMRRConv module}

\textit{Internal components}. We examine the roles of local transition responses, channel-axis propagation, and scale diversity through four variants of AMRRConv. The variants remove the Laplacian-initialized branch, remove the Sobel-initialized branch, disable channel-axis propagation, or restrict all multi-scale kernels to $k=3$. Fig.~\ref{fig:component_structure} compares the full model and these variants using descriptors from the same image patch to assess local contrast and texture reconstruction against the reference.

Local contrast (LC) characterizes local intensity variation, local binary pattern (LBP) encodes neighbourhood texture patterns, and multi-scale Gaussian detail (MSGD) captures detail responses from fine to coarse scales. For the common image patch in Fig.~\ref{fig:component_structure}, the discrepancy from the reference is the mean $L_1$ distance between corresponding normalized descriptor vectors, defined as
\begin{equation}
D_{\mathrm{feat}}(I)=\frac{1}{N}\sum_p\left\|
\hat{\phi}_{I}(p)-\hat{\phi}_{\mathrm{GT}}(p)
\right\|_1,
\label{eq:component_structure_distance}
\end{equation}
where $N$ is the number of pixels in the common window and $\hat{\phi}$ denotes the normalized descriptor vector.
The descriptor vector is defined as
\begin{equation}
\phi=[\mathrm{LC},\mathrm{LBP},\mathrm{MSGD}].
\label{eq:component_structure_descriptor}
\end{equation}

The full model achieves the lowest $D_{\mathrm{feat}}$ of 0.449 in the selected patch. Removing the Laplacian-initialized branch increases the error to 0.473 and produces smoother weak patterns, while removing the Sobel-initialized branch raises it to 0.514 and leaves less distinct boundaries. Disabling channel-axis propagation increases the error to 0.472, and restricting all multi-scale kernels to $3\times3$ gives 0.503. These comparisons support combining local transition responses, channel-axis propagation, and multiple neighbourhood scales for local contrast and texture reconstruction.

\textit{Propagation form}. Table~\ref{tbl:propagation} compares spatial-only, joint-3D, and tri-planar propagation with the same budget of $3k^2$ kernel coefficients per scale. The response branches, fusion modules, and backbone remain unchanged. Tri-planar propagation gives the best scores on all five measures, reaching 21.84\,dB PSNR and 0.8080 SSIM. Relative to Joint-3D, PSNR improves by approximately 0.63\,dB and luminance MAE decreases from 0.0760 to 0.0656. These results support the proposed organization of local feature interactions along the channel and spatial axes in AMRRConv.
\begin{table}[pos=!htbp]
\centering
\caption{Propagation operators on the LOLv1 test set, with a $3k^2$ kernel-coefficient budget per scale. Deep orange and light orange mark the best and second-best scores.}\label{tbl:propagation}
\fontsize{7}{8.5}\selectfont
\setlength{\tabcolsep}{1.5pt}
\renewcommand{\arraystretch}{1.15}
\begin{tabular*}{\columnwidth}{@{\extracolsep{\fill}}llrrrrr@{}}
\toprule
\makecell[l]{Propagation\\operator} & \makecell[l]{Local\\support} & PSNR$\uparrow$ & SSIM$\uparrow$ & GMSD$\downarrow$ & \makecell[r]{CIEDE2000\\$\downarrow$} & \makecell[r]{Luminance\\MAE$\downarrow$} \\
\midrule
Spatial-2D & $3\times(1\times k\times k)$ & 18.4154 & 0.7347 & 0.1072 & 13.1748 & 0.1072 \\
Joint-3D & $3\times k\times k$ & \seco{21.2144} & \seco{0.8028} & \seco{0.0694} & \seco{9.2758} & \seco{0.0760} \\
\makecell[l]{Tri-planar\\(ours)} & \makecell[l]{$C\!\text{-}\!H,\ C\!\text{-}\!W,$\\$H\!\text{-}\!W$} & \best{21.84} & \best{0.8080} & \best{0.0632} & \best{8.5963} & \best{0.0656} \\
\bottomrule
\end{tabular*}
\end{table}

\textit{Complete module replacement}. Table~\ref{tbl:module_effect} compares AMRRConv with a $1\times1$ projection, ConvNeXt~\cite{liu2022convnext}, SCConv~\cite{liu2020scconv}, and InceptionNeXt~\cite{yu2024inceptionnext} at brightness adjustment map prediction and residual feature fusion. AMRRConv achieves the highest PSNR and SSIM and the lowest GMSD, CIEDE2000, and luminance MAE among the tested modules. Relative to the projection baseline, it improves PSNR by approximately 2.04\,dB and SSIM by 0.0153, while reducing luminance MAE from 0.0851 to 0.0656. It also exceeds InceptionNeXt, the strongest alternative in PSNR, by 0.42\,dB PSNR and 0.0072 SSIM.
\begin{table}[pos=!htbp]
\centering
\caption{Module replacement on the LOLv1 test set. State elements include learned parameters and stored running statistics at both replacement locations. Deep orange and light orange mark the best and second-best scores.}\label{tbl:module_effect}
\fontsize{7}{8.5}\selectfont
\setlength{\tabcolsep}{1.5pt}
\renewcommand{\arraystretch}{1.15}
\begin{tabular*}{\columnwidth}{@{\extracolsep{\fill}}lrrrrrr@{}}
\toprule
Module & \makecell[r]{State\\(K)} & PSNR$\uparrow$ & SSIM$\uparrow$ & GMSD$\downarrow$ & \makecell[r]{CIEDE2000\\$\downarrow$} & \makecell[r]{Luminance\\MAE$\downarrow$} \\
\midrule
\makecell[l]{w/o AMRRConv\\($1\times1$ projection)} & 0.17 & 19.8015 & 0.7927 & \seco{0.0664} & 10.2592 & 0.0851 \\
ConvNeXt & 10.34 & 20.12 & 0.7731 & 0.0809 & 9.9825 & 0.0847 \\
SCConv & 1.43 & 20.76 & 0.7960 & 0.0687 & 9.2457 & \seco{0.0762} \\
InceptionNeXt & 0.99 & \seco{21.42} & \seco{0.8008} & 0.0763 & \seco{9.1737} & 0.0770 \\
AMRRConv (ours) & 33.99 & \best{21.84} & \best{0.8080} & \best{0.0632} & \best{8.5963} & \best{0.0656} \\
\bottomrule
\end{tabular*}
\end{table}

In Fig.~\ref{fig:module_visual}, AMRRConv reduces pronounced residual errors near cabinet edges and object contours compared with the projection baseline. The gradient-difference map also shows reduced mismatch along several local boundaries. Row~\textbf{D} maps errors in local correlations between RGB gradient magnitudes, complementing the pixel and gradient comparisons with a measure of inter-channel structural consistency in reconstruction.

\subsubsection{Guidance Cues}
Table~\ref{tbl:guidance_cues} evaluates guidance-cue removal while retaining AMRRConv and both reconstruction paths. Removing chroma or shallow texture lowers PSNR by approximately 1.39\,dB and 1.27\,dB, respectively, and worsens SSIM, colour error, and luminance error. Removing all guidance cues reduces PSNR to 21.0170\,dB and SSIM to 0.7892. The DoG-based luminance-variation cue provides a smaller PSNR gain. Its removal slightly improves SSIM and GMSD but increases CIEDE2000 from 8.5963 to 9.3047 and luminance MAE from 0.0656 to 0.0705.

\begin{table}[pos=!htbp]
\centering
\caption{Guidance-cue ablation on the LOLv1 test set. Deep orange and light orange mark the best and second-best scores.}\label{tbl:guidance_cues}
\fontsize{7}{8.5}\selectfont
\setlength{\tabcolsep}{1.5pt}
\renewcommand{\arraystretch}{1.15}
\begin{tabular*}{\columnwidth}{@{\extracolsep{\fill}}lrrrrr@{}}
\toprule
Configuration & PSNR$\uparrow$ & SSIM$\uparrow$ & GMSD$\downarrow$ & \makecell[r]{CIEDE2000\\$\downarrow$} & \makecell[r]{Luminance\\MAE$\downarrow$} \\
\midrule
Full (ours) & \best{21.84} & \seco{0.8080} & \seco{0.0632} & \best{8.5963} & \best{0.0656} \\
w/o DoG cue & \seco{21.7547} & \best{0.8107} & \best{0.0616} & \seco{9.3047} & \seco{0.0705} \\
w/o chroma cue & 20.4545 & 0.7974 & 0.0693 & 9.7323 & 0.0856 \\
w/o texture cue & 20.5661 & 0.7864 & 0.0657 & 9.9331 & 0.0815 \\
\makecell[l]{w/o all guidance\\cues} & 21.0170 & 0.7892 & 0.0730 & 9.8231 & 0.0799 \\
\bottomrule
\end{tabular*}
\end{table}

Fig.~\ref{fig:ole_guidance} complements the full-image scores with matched colour and texture patches. Full gives the lowest CIEDE2000 of 14.48 in the orange patch and the lowest gradient error of 0.0248 in the cyan patch. Removing texture increases both errors to 17.75 and 0.0262, while removing luminance variation gives the largest gradient error of 0.0265 in this patch. Removing chroma increases colour error to 15.21, while its gradient error of 0.0249 remains close to Full. These observations illustrate the contribution of the cues to colour fidelity and local pattern reconstruction.

\subsubsection{Reconstruction Paths}
Table~\ref{tbl:dual_path} compares reconstruction using only the brightness adjustment path, only the residual path, or both paths, with all three guidance cues retained. Residual-only provides the stronger single-path result. Combining both paths increases PSNR by approximately 0.84\,dB and SSIM by 0.0078 over residual-only, reaching 21.84\,dB and 0.8080. It also reduces luminance MAE from 0.0736 to 0.0656 and achieves the lowest GMSD and CIEDE2000 among the three variants. These improvements support combining iterative photometric adjustment with feature-based reconstruction of local content.

\begin{table}[pos=!htbp]
\centering
\caption{Reconstruction-path ablation on the LOLv1 test set. Deep orange and light orange mark the best and second-best scores.}\label{tbl:dual_path}
\fontsize{7}{8.5}\selectfont
\setlength{\tabcolsep}{1.5pt}
\renewcommand{\arraystretch}{1.15}
\begin{tabular*}{\columnwidth}{@{\extracolsep{\fill}}lrrrrr@{}}
\toprule
Configuration & PSNR$\uparrow$ & SSIM$\uparrow$ & GMSD$\downarrow$ & \makecell[r]{CIEDE2000\\$\downarrow$} & \makecell[r]{Luminance\\MAE$\downarrow$} \\
\midrule
\makecell[l]{Brightness\\adjustment path\\only} & 15.8017 & 0.5139 & 0.0949 & 16.6962 & 0.1445 \\
Residual path only & \seco{21.0046} & \seco{0.8002} & \seco{0.0663} & \seco{8.9849} & \seco{0.0736} \\
Full (ours) & \best{21.84} & \best{0.8080} & \best{0.0632} & \best{8.5963} & \best{0.0656} \\
\bottomrule
\end{tabular*}
\end{table}

Figs.~\ref{fig:ole_paths} and \ref{fig:ole_profile} examine the common hanger region and its luminance profile. Using only the brightness adjustment path overestimates brightness across several intervals, while residual-only leaves the corresponding bright regions below the reference. Combining the paths reduces mean patch luminance error to 0.047, compared with 0.171 for the brightness adjustment path alone and 0.089 for residual-only. Full also brings several bright intervals closer to the reference along the shared scanline, improving local intensity levels while preserving the repeated transitions.

\subsection{AMRRConv Transfer to Low-Light Object Detection}\label{sec:downstream}
We assess AMRRConv in low-light object detection by replacing two YOLOv11 neck blocks at feature resolutions of $80\times80$ and $40\times40$. AMRRConv is compared with NAF~\cite{chen2022naf}, ContMix~\cite{lou2025overlock}, Restormer~\cite{zamir2022restormer}, RCAB~\cite{zhang2018rcan}, ConvAttn, and MSRB~\cite{li2018multi}. All variants are trained on ExDark~\cite{loh2019exdark} for 300 epochs using SGD and $640\times640$ inputs. Fig.~\ref{fig:yolo} compares detection performance, neck parameter counts, and operation estimates across the tested modules.

AMRRConv achieves a precision of 67.71\%, ranking third among the seven modules. Across the four detection metrics, its largest gap from the best score is only 1.34 percentage points, in mAP$_{50}$ (53.50\% versus 54.84\%). It uses the fewest neck parameters at 0.115\,M, approximately $1/31$ of MSRB's parameter count, with an estimated neck cost of 0.500\,GFLOPs.

\FloatBarrier
\section{Conclusion}\label{sec:conclusion}
We present CueDR-Net, which combines brightness adjustment with local content recovery through bidirectional interaction between image features and auxiliary cues. AMRRConv aggregates responses across scales according to local content to refine features for both the brightness adjustment path and the residual path. On LOLv1, LSRW-HUAWEI, and LSRW-NIKON, CueDR-Net achieves the highest PSNR among the compared methods with competitive structural similarity. Ablation studies further support the effectiveness of auxiliary cues, combined reconstruction paths, and AMRRConv. In ExDark detection experiments, AMRRConv delivers competitive performance with the fewest neck parameters among the tested modules, demonstrating its potential for use in low-light object detection networks.

\section*{Acknowledgments}
This work was supported in part by the National Key Research and Development Program of China under Grant 2025YFB2606504.

\section*{CRediT authorship contribution statement}
\textbf{Yicui Shi:} Conceptualization, Methodology, Visualization, Writing - original draft, Writing - review \& editing, Resources, Validation.
\textbf{Yuhan Chen:} Data curation, Investigation, Writing - review \& editing.
\textbf{Xiangfei Huang:} Validation, Visualization.
\textbf{Zhenguo Wang:} Visualization, Resources.
\textbf{Wenxuan Yu:} Validation, Investigation.
\textbf{Ying Fang:} Validation, Investigation.
\textbf{Jie Li:} Conceptualization, Investigation, Writing - review \& editing, Methodology.
\textbf{Guofa Li:} Conceptualization, Writing - review \& editing.

\section*{Declaration of Competing Interest}
The authors declare that they have no known competing financial interests or personal relationships that could have appeared to influence the work reported in this paper.

\section*{Data availability}
The LOLv1, LSRW, and ExDark datasets used in this study are available from their respective providers~\cite{wei2018deep,hai2023r2rnet,loh2019exdark}.

\FloatBarrier
\bibliographystyle{elsarticle-num-plain} 

\bibliography{cas-refs}

\end{document}